%% file: main.tex
\documentclass{article}
\usepackage[total={6in, 8in}]{geometry}
\usepackage{amsmath}
\usepackage{amsfonts}
\usepackage{graphicx}
\usepackage{epstopdf}
\usepackage{algorithmic}
\usepackage{mathrsfs}
\usepackage{booktabs}
\usepackage{hyperref}
\usepackage{cleveref}
\usepackage[caption=false]{subfig}
\usepackage{tikz-cd}
\ifpdf
  \DeclareGraphicsExtensions{.eps,.pdf,.png,.jpg}
\else
  \DeclareGraphicsExtensions{.eps}
\fi

\newcommand{\bfe}{\mathbf{e}}

\newcommand{\baralpha}{\bar{\alpha}}
\newcommand{\barw}{\overline{W}}
\newcommand{\bary}{\overline{Y}}

\newcommand{\N}{\mathbb{N}}
\newcommand{\R}{\mathbb{R}}
\newcommand{\T}{\mathbb{T}}

\newcommand{\Lm}{\mathcal{L}}
\newcommand{\mcM}{\mathcal{M}}
\newcommand{\mcN}{\mathcal{N}}

\newcommand{\msC}{\mathscr{C}}
\newcommand{\msD}{\mathscr{D}}
\newcommand{\msL}{\mathscr{L}}

\newcommand{\floor}[1]{\left\lfloor #1 \right\rfloor}

\newcommand{\wone}{\overline{W}_1}
\newcommand{\yone}{\overline{Y}_1}

\newcommand{\wtwo}{\overline{W}_2}
\newcommand{\ytwo}{\overline{Y}_2}

\newcommand{\hn}[1]{\mathcal{H}^{#1}}
\newcommand{\hl}{\mathcal{H}^{\mathcal{L}}}

\newcommand{\ra}[1]{\renewcommand{\arraystretch}{#1}}

\usepackage{enumitem}
\setlist[enumerate]{leftmargin=.5in}
\setlist[itemize]{leftmargin=.5in}

\title{The SVD of Convolutional Weights: A CNN Interpretability Framework\thanks{The research described in this paper is part of the MARS
Initiative at Pacific Northwest National Laboratory. It
was conducted under the Laboratory Directed Research
and Development Program at PNNL, a Multiprogram
National Laboratory operated by Battelle Memorial
Institute for the U.S. Department of Energy under Contract
DE-AC05-76RL01830.}}

\author{Brenda Praggastis\thanks{AI \& Data Analytics Division, Pacific Northwest National Laboratory, Seattle, WA 98109.}
\and Davis Brown\footnotemark[2]
\and Carlos Ortiz Marrero\thanks{AI \& Data Analytics Division, Pacific Northwest National Laboratory, Richland, WA 99354; Department of Electrical \& Computer Engineering, North Carolina State University, Raleigh, NC 27607.}
\and Emilie Purvine\footnotemark[2]
\and Madelyn Shapiro\footnotemark[2]
\and Bei Wang\thanks{Scientific Computing and Imaging (SCI) Institute, University of Utah, Salt Lake City, UT 94112.}}

\usepackage{amsopn}
\DeclareMathOperator*{\argmax}{arg\,max}

\begin{document}

\maketitle

\begin{abstract}
\input{abstract}

\end{abstract}

\section{Introduction}
\label{sec:introduction}
\input{introduction}

\section{Motivation and Background}
\label{sec:back}
\input{background}

\section{Preliminaries}
\label{sec:preliminaries}
\input{preliminaries}

\section{The SVD in the Convolutional Layers}
\label{sec:svd}
\input{svdofw}

\section{Hypergraphs and the Model’s Semantic Hierarchy}
\label{sec:semhh}
\input{semhh}

\section{A Suggested Framework for Interpretability}
\label{sec:framework}
\input{framework}

\section{Conclusion}
\label{sec:conclusion}
\input{conclusion}

\bibliographystyle{plain}
\bibliography{references}

\end{document}

%% file: abstract.tex
Deep neural networks used for image classification often use convolutional filters to extract distinguishing features before passing them to a linear classifier.
Most interpretability literature focuses on providing semantic meaning to convolutional filters to explain a model's reasoning process and confirm its use of relevant information from the input domain.
Fully connected layers can be studied by decomposing their weight matrices using a singular value decomposition, in effect studying the correlations between the rows in each matrix to discover the dynamics of the map.
In this work we define a singular value decomposition for the weight tensor of a convolutional layer, which provides an analogous understanding of the correlations between filters, exposing the dynamics of the convolutional map. 
We validate our definition using recent results in random matrix theory.
By applying the decomposition across the linear layers of an image classification network we suggest a framework against which interpretability methods might be applied using hypergraphs to model class separation. 
Rather than looking to the activations to explain the network, we use the singular vectors with the greatest corresponding singular values for each linear layer to identify those features most important to the network.
We illustrate our approach with examples and introduce the DeepDataProfiler library, the analysis tool used for this study.

%% file: introduction.tex
Mathematical functions and equations provide elegant and concise expression of the relationships and dynamics of physical systems.
While we might not understand the derivation or full significance of all the parameters in a given equation, we can still be persuaded to rely on its predictive value.
We can be shown how to interpret the equation by linking its parameters to important values in the system and by expressing their relationships in terms of the dynamics of the system.
Machine learning practitioners have long striven to obtain this same kind of interpretability for the trained neural networks they produce, but have had limited success due to their size and complexity \cite{dey2022human,fan2020interpretability,rudin2021interpretable}.

A trained deep learning model defines a system of multivariate continuous functions or tensor maps composed of linear and nonlinear maps referenced as layers.
The success or failure of the model depends largely on the parameters or weights found in the linear layers, which are defined during a model's training process.
Consequentially, much interpretability literature offers methods for discovering semantic meaning from the feature maps or activations produced by the linear layers in order to tie them to the input domain  \cite{olah2018building,bau2017network,hohman2020summit}.

Convolutional neural networks (CNNs) used for image classification provide the most accessible opportunities to derive semantic meaning from feature maps because the features of interest are visual objects.
Image tensors can be used to probe network response by measuring their activation values \cite{olah2017feature}.
Strong responses in convolutional layers indicate strong correlation (positive or negative) to the filters in the weight tensor,
and in fully connected layers to the rows of the weight matrix.
But, what does a strong correlation mean?

The singular value decomposition (SVD) holds a special place in the heart of data science and numerical linear algebra \cite{stewart1993early}. As a robust matrix factorization method it facilitates data compression \cite{wei2001ecg,andrews1976svd} and network pruning \cite{psichogios1994svdnet}.
For our purpose the unitary nature of the singular vectors make the SVD an ideal factorization for understanding the dynamics, and hence the correlations measured by the weight tensors.

In the case of a multi-layer perceptron,
 domain concepts producing activations highly correlated to a singular vector are scaled by the corresponding singular value.
Our belief is that these basic correlations drive the success of the model so that interpretability rests largely on learning what input features correlate most closely to the singular vectors.
We define an SVD for convolutional layers in terms of the SVD for an unfolding of the weight tensor. This echos the handling of convolutional layers in Yoshida and Miyato \cite{yoshida2017spectral} but differs from the approach used by Sedghi, Gupta, and Long \cite{sedghi2019singular}.
We describe the differences in these approaches and
demonstrate that our definition is consistent with notional ideas around correlation as well as
recent empirical results on the spectral distribution of covariance matrices associated with fully connected layers.

Our contribution brings a novel perspective to interpretability research.
Much of the work of feature attribution  and visualization is focused on the channel activations generated by convolutional layers \cite{olah2018building,hohman2020summit,olah2017feature,zhang2018interpreting,olah2020overview}.
We consider the linear dynamics of the convolutional layers as transformations restricted to the subspace of receptive fields, the regions acted on by convolution.
By shifting to the receptive field space it is possible to unfold the weight tensors into matrices to study their dynamics.
This process gives mathematical meaning to the parameters and opens the door to discovery of unanticipated features in  the input domain.

The paper is organized in two parts.
\Cref{sec:back,sec:preliminaries,sec:svd} give the mathematical justification behind our use of the SVD for convolutional layers.
\Cref{sec:semhh,sec:framework} describe how the SVD might be used to guide interpretability research.
We close with examples, and introduce the DeepDataProfiler library \cite{deepdataprofiler} used for this work.

%% file: background.tex
The singular value decomposition of a real-valued matrix $M$ is a factorization  $M = USV^T $ into two orthonormal change of basis matrices $U$ and $V$, whose columns are called singular vectors, and a diagonal matrix $S$ of singular values.
When viewed as a linear transformation, $M:\R^m \rightarrow \R^n$, this decomposition exposes the dynamical behavior of the transformation as a scaling of the subspaces spanned by the singular vectors \cite{brunton2022data}.
The eigenvalues of the correlation matrix $MM^T$ are the squares of the singular values.

Applying random matrix theory to deep learning, Martin and Mahoney \cite{martin2020heavytailed,martin2021implicit,martin2021predicting} examine the distribution of the eigenvalues of $MM^T$ when $M$ is the   weight matrix of a fully connected layer of a trained neural network.
They use these distributions to define a capacity metric for the model, which correlates with the quality of the training process.
Their work points to the use of the SVD to understand training dynamics and the potential importance of the singular values in understanding the decision process used by the model.
We discuss this more in \cref{sec:svd}.

In their study of the learning dynamics of neural networks, Saxe, McClelland, and Ganguli \cite{saxe2019mathematical} demonstrate that the training of a shallow multi-layer perceptron involves learning the semantic hierarchical structure of the domain data. Moreover, this knowledge is captured by the singular vectors and singular values learned by the network.
This work is notable for its use of the projection onto the singular vectors as a measure of the importance of a feature for an individual classification.

While many different projections of hidden layer activations appear to be semantically coherent \cite{szegedy2014intriguing}, Bau et al. \cite{bau2017network} find evidence that the representations closer to the Euclidean (or ``natural'') basis are more meaningful than random unitary transformations. This makes sense in part due to the element-wise nature of the activation maps following the linear transformations.
But the dynamics of the linear transformations were learned under the constraints of the model's architecture.
For important domain features to persist as they are passed through the network, they must be scaled to persist through regularization.
In particular, Dittmer, King, and Maass \cite{dittmer2020singular} demonstrate that only activations aligned to singular vectors with the largest corresponding singular values persist past ReLU.

From the above observations we infer that not only do the singular values indicate quality of training, but the singular vectors themselves may hold the key to understanding the latent features of the model.
For this reason we use the singular vectors of an SVD to study the influence of convolutional and fully connected linear transformations on the network.

Both Saxe, McClelland, and Ganguli \cite{saxe2019mathematical} and Martin and Mahoney \cite{martin2020heavytailed,martin2021implicit,martin2021predicting} restrict their analysis to the fully connected linear layers.\footnote{Martin and Mahoney \cite{martin2020heavytailed,martin2021implicit,martin2021predicting} address convolutional layers but do not flesh out the details.} To extend their work to all layers of a CNN, our first task is to extend the singular value decomposition to the convolutional layers and project the layer-wise activations onto the singular vectors to measure correlations.

A common approach for interpreting the activations produced by the weights of the hidden layers is to optimize an input to the CNN that reliably produces a large response in an activation of interest.
This method has been most notably used to create \textit{feature visualizations}, images optimizing a feature map, for image classification networks \cite{szegedy2014intriguing, mahendran2015understanding, wei2015understanding, nguyen2016multifaceted}.
We draw on these optimization techniques to gain an understanding of the concepts in the domain that correlate with the singular vectors for a set of examples.

Interpretability research often depends upon linking feature maps to predetermined human identifiable concepts \cite{bau2017network,ribeiro2016why,kim2018interpretability}.
For example, Concept Activation Vectors \cite{kim2018interpretability}  and LIME \cite{ribeiro2016why} start with human-defined concepts and measure the model's sensitivity to them. 
These are powerful tools for verifying the model's ability to recognize important domain-centric concepts.
But interpretability research starting from the premise that latent representations of trustworthy models must correspond to known domain-centric concepts assumes we know in advance everything in the domain that is meaningful.
This could produce bias and eliminate the possibility that concepts used by the model to describe class distinctions could be very different than what we expect and yet still be domain-centric and legitimate for classification.
As a consequence we will not try to incorporate these approaches as we are interested in first discovering what the model defines as important, and only then attaching them to something human interpretable.
The subtle difference in discovering the semantics of the network versus the sensitivity of the network to predefined concepts is discussed extensively in Olah et al. \cite{olah2018building,olah2020zoom}.

%% file: preliminaries.tex
Let  $\mathscr{D}$ be a domain of images partitioned into target classes $\msC = \{C^j\}_{j=1}^N$.
Let $\mcM$ be an image classification CNN  trained and tested to classify the images in  $\mathscr{D}$ into its $N$ target classes.
$\mcM$ is composed of of multiple \emph{tensor maps} called \emph{layers}.
As discussed in the introductory remarks, our focus will be on a network's linear layers (fully connected and convolutional)\footnote{A linear layer usually references a linear map followed by addition of a bias term and some non-linear map.  For this work we will mean only the linear map when we reference a linear layer.}, which we denote as $\mathscr{L}$.
Other layers in the network, such as ReLU, batch normalization, and pooling, are fixed by the architecture and learn any needed parameters indirectly from the layers in $\mathscr{L}$.
While these other layers correspond to architectural considerations designed to constrain the learning process during training, improving accuracy and generalizability, the latent representations of the input data used for classification are ultimately determined by the linear layers.

The SVD expresses a linear map between vector spaces as a weighted sum of singular values times singular vectors.
The SVD is commonly used for low rank approximations in data science \cite{brunton2022data}.
We are interested in the geometric properties of a linear map exposed by the SVD, which describe the dynamics of the mapping.

The computation of the SVD of weight tensors in the fully connected layers is straightforward.
However, while the linearity properties of convolution are well-understood, there is more than one way to express this operation using 2-tensors or matrices.
The choice of representation affects the computational complexity of generating the SVD, and more importantly,  informs the perspective from which we extract meaning from the matrix.
Our goal is to choose a representation that accurately reflects the role the operation plays in a neural network and provides us with informative singular values within this context.

\subsection{Cross-Correlation as Matrix Multiplication}
\label{subsection:multiplication}

First we recognize that the convolution used in a CNN is really cross-correlation, also known as a sliding dot product.
Let $\Lm$ be a convolutional layer in $\msL$ with weight tensor $W$. Let $X$ be an input tensor to $\Lm$ and  $Y=W \star X$, where $\star$ means cross-correlation.
For ease of notation and without loss of generality, we make some basic assumptions about $W$ and $X$.
We assume that $X$ is a $3$-tensor of dimensions $c \times m \times m$,
$W$ is a $4$-tensor of dimensions $d \times c \times k \times k$ where $k \leq m$, the stride is $1$, and the operation is $2$-dimensional.

Let $I = (i_0,i_1,\dots,i_{s-1}) \in \N^s$ denote a tuple.
Let $\T_{I}$ be the set of real-valued $s$-tensors (i.e., $s$-order tensors) where $I$ indicates the size of each of the $s$ dimensions.
Each tensor $T \in \T_I$ is indexed by the set
$$\text{Index}(\T_I) = \{\alpha \in \N^s: \alpha_j < i_j, \forall j < s\},$$ and follows a row-major ordering.
Following these conventions we say $W \in \T_{(d,c,k,k)}$ has $d$ \textit{filters}, each  a  $3$-tensor with $c$ \textit{channels} and \textit{spatial dimensions} of $k \times k$.
Similarly, we say $X \in \T_{(c,m,m)}$ has $c$ channels and spatial dimensions $m \times m$.
In this setting, $Y \in \T_{(d,n,n)}$, for $n = m-k+1$, has $d$ channels and spatial dimensions $n \times n$. An element of $Y$ is a valid\footnote{By valid we mean that every term in the sum exists. Any padding required by the model to preserve tensor size is assumed to have already been applied to $X$.}
sum of products:
\begin{equation}\label{equation:yewx1}
Y_{h,i,j} =  (W \star X)_{h,i,j}  = \sum_{r,s,t} W_{h,r,s,t} \cdot X_{r,i+s,j+t}.
\end{equation}
The \textit{channel activations} of $Y$ are its 2-dimensional slices  indexed by the spatial dimensions and referenced as $Y_r = Y[r,:,:]$.
The \textit{spatial activations} are its $1$-dimensional slices indexed by the channels and referenced as $Y_{i,j} = Y[:,i,j]$.

The set $\T_I$ along with element-wise addition and scalar multiplication is a real-valued vector space isomorphic to $\R^{\Pi I}$, where $\Pi I = \Pi_{k=0}^{s-1} i_k$ is the dimension of the vector space, i.e., the cardinality of its basis.
As we will see, it is helpful to have access to this isomorphism.
Let $\{\bar{\bfe}_{\bar{\alpha}}\}_{\bar{\alpha} \in [0..\Pi I - 1]}$ be the standard Euclidean basis for $\R^{\Pi I}$.
Define the corresponding Euclidean basis for $\T_I$ using the subset of $\T_I$ given by  $\{\bfe^I_\alpha\}_{\alpha \in \text{Index}(\T_I)}$ such that $(\bfe^I_{\alpha})_{\beta} = \delta_{\alpha\beta}$\footnote{The notation means that  the $\beta^{\text{th}}$ component of the $\alpha^{\text{th}}$ basis vector is equal to the Kronecker delta: $\delta_{\alpha\beta} = 1$ if $\alpha = \beta$ and $0$ otherwise.} for $\beta \in \text{Index}(\T_I)$.

Define an isomorphism
$\phi_I: \T_I \rightarrow \R^{\Pi I}$ by pairing the basis vectors using their indices: order the indices in Index($\T_I$) in ascending lexicographic order and pair each index with its position in the list.
It is straightforward to check, if $\phi_I(\bfe_\alpha) = \bar{\bfe}_{\baralpha}$ then
\begin{equation}\label{equation:reshape1}
    \baralpha = \sum_{j=0}^{s-2} \left[\alpha_j \cdot \Pi_{\beta = j+1}^{s-1} i_\beta\right] + \alpha_{s-1}.
\end{equation}
For any tensor $T \in \T_I$, $\phi_I(T)$ is the \textit{vectorized} form of $T$. We will often refer to this as the \textit{flattening map}.
If $\phi_I^{-1}(\bar{\bfe}_{\baralpha}) = \bfe_\alpha$, then $\alpha$ is defined recursively using the formulas:
\begin{equation}\label{equation:reshape2}
    \begin{split}
        \alpha_0 & = \floor{\frac{\baralpha}{\Pi_{\beta=1}^{s-1} i_\beta}} \\
        \alpha_j & = \floor{\frac{\baralpha - \sum_{\gamma=0}^{j-1} \alpha_\gamma \cdot \Pi_{\beta=\gamma+1}^{s-1} i_\beta}{\Pi_{\beta=j+1}^{s-1} i_\beta}} \text{for } j<s-1  \\
        \alpha_{s-1} & = \baralpha - \sum_{\gamma=0}^{s-2} \alpha_\gamma \cdot \Pi_{\beta=\gamma+1}^{s-1} i_\beta.
    \end{split}
\end{equation}
This isomorphism is useful because it provides a consistent way to reshape a tensor.
For suppose $J \in \N^{s'}$ such that $\Pi J = \Pi I$, then $\phi_{IJ} = \phi_J^{-1} \circ \phi_I : \T_I \rightarrow \T_J$ is also an isomorphism.
In general we call $\phi_I$, $\phi_J$ and $\phi_{IJ}$ \textit{reshaping maps} and reference them simply as $\phi$ when $I$ and/or $J$ are clear from context.

In order to rewrite \cref{equation:yewx1} as a simple dot product of two $1$-dimensional tensors,
define a projection map $\psi^{(m,k)}_{i,j}:\T_{(c,m,m)} \rightarrow \T_{(c,k,k)}$ as
$$\psi^{(m,k)}_{i,j}(X) = \sum_{r\leq c;s,t \leq k} X_{r,i+s,j+t}\cdot \bfe_{r,s,t}.$$
The tensor $\psi^{(m,k)}_{i,j}(X) \in \T_{(c,k,k)}$ is the \textit{receptive field} associated with the spatial
activation vector in $Y$ with spatial index $(i,j)$.
Now \cref{equation:yewx1} becomes:
\begin{equation}\label{equation:yewx2}
    Y_{h,i,j} = \phi_{(c,k,k)}(W_{h}) \cdot \phi_{(c,k,k)}\circ\psi^{(m,k)}_{i,j}(X).
\end{equation}
The first term in the dot product is the flattened filter $W_h$. The second term in the dot product is a flattened receptive field in $X$.

The inverse of the projection provides a natural embedding
$(\psi^{(m,k)}_{i,j})^{-1}:\T_{(c,k,k)}\rightarrow\T_{(c,m,m)}$,
such that for all $T \in \T_{(c,k,k)}$:
$$(\psi_{i,j}^{(m,k)})^{-1}(T) = \sum_{r,s,t} T_{r,s,t} \cdot \bfe_{r,i+s,j+t}.$$
Use the embedding map to rewrite \cref{equation:yewx1} as:
\begin{equation}\label{equation:yewx3}
    Y_{h,i,j} = \phi_{(c,m,m)}\circ(\psi^{(m,k)}_{i,j})^{-1}(W_{h}) \cdot \phi_{(c,m,m)}(X).
\end{equation}
The first term in this dot product is the filter $W_h$ embedded in $\T_{(c,m,m)}$ and then flattened into $\R^{cm^2}$.
The second term in the dot product is the flattened input tensor $X$.
The difference between \cref{equation:yewx2,equation:yewx3} is the domain where the dot product is being performed.
In \cref{equation:yewx2} the vectors belong to $\R^{ck^2}$, while in \cref{equation:yewx3} the vectors belong to $\R^{cm^2}$.
Typically $k<<m$ so that \cref{equation:yewx2} would seem preferable.

\subsection{Two Matrix Representations of Cross-Correlation}
\label{subsection:twomatrix}

 Cross-correlation can be represented  as the bilinear map: 
\begin{equation}\label{equation:chi}
    \chi : \T_{(d,c,k,k)} \times \T_{(c,m,m)} \rightarrow \T_{(d,n,n)}.
\end{equation}
The bilinear map in turn can be expressed as matrix multiplication using either of
\cref{equation:yewx2} or \cref{equation:yewx3}.

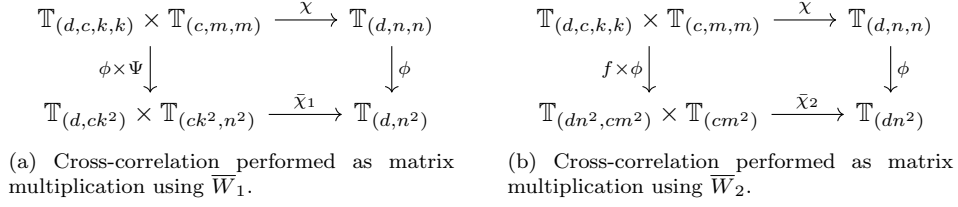
\begin{figure}[t]
	\centering
	\subfloat[Cross-correlation performed as matrix multiplication using $\wone$.]{
		\label{fig:cdiag1}
		\begin{tikzcd}[ampersand replacement=\&]
			\T_{(d,c,k,k)} \times \T_{(c,m,m)} \arrow[r, "\chi"] \arrow[d, "\phi \times \Psi" swap]
			\& \T_{(d,n,n)} \arrow[d, "\phi"] \\
			\T_{(d,ck^2)} \times \T_{(ck^2,n^2)} \arrow[r, "\bar{\chi}_1"]
			\& \T_{(d,n^2)}
		\end{tikzcd}
	}
	\qquad
	\subfloat[Cross-correlation performed as matrix multiplication using $\wtwo$.]{
		\label{fig:cdiag2}
		\begin{tikzcd}[ampersand replacement=\&]
			\T_{(d,c,k,k)} \times \T_{(c,m,m)} \arrow[r, "\chi"] \arrow[d, "f \times \phi" swap]
			\& \T_{(d,n, n)} \arrow[d, "\phi"] \\
			\T_{(dn^2,cm^2)} \times \T_{(cm^2)} \arrow[r, "\bar{\chi}_2"]
			\& \T_{(dn^2)}
		\end{tikzcd}
	}
	\caption{Matrization of weight tensors.}
\end{figure}

\Cref{equation:yewx2} performs the dot product in $\R^{ck^2}$.
We reshape $W\in\T_{(d,c,k,k)}$ into  $\wone = \phi(W) \in \T_{(d,ck^2)}$, so that $\wone$ is a matrix, where each row is a flattened filter from $W$.
Note that $\wone$ is an \textit{unfolding} of the weight tensor similar to what is described in Kolda and Bader \cite{kolda2009tensor}.
We also reshape $Y \in \T_{(d,n,n)}$ to $\yone = \phi(Y) \in \T_{(d,n^2)}$.
Define the receptive field matrix $\Psi(X) \in \T_{(ck^2,n^2)}$ so that each column is the flattened receptive field of $X$ given by
\begin{equation}
    \Psi(X)_{:,i \cdot n+j} = (\phi_{(c,k,k)}\circ\psi^{(m,k)}_{i,j}(X))^T.
\end{equation}
We now represent the cross-correlation  $W \star X$ as the matrix multiplication:
\begin{equation}
    \wone\cdot\Psi(X) = \yone.
    \label{equ:m1rep}
\end{equation}
Let $\bar{\chi}_1$ be the bilinear map defined by \cref{equ:m1rep} required to complete the commutative diagram in \cref{fig:cdiag1}.

\begin{figure}[t]
	\centering
	\subfloat[$\wone$ is a reshaped weight tensor.]{
	    \label{fig:matrep1}
	    \includegraphics[width=0.25\linewidth,trim=0 0 0 0,clip]{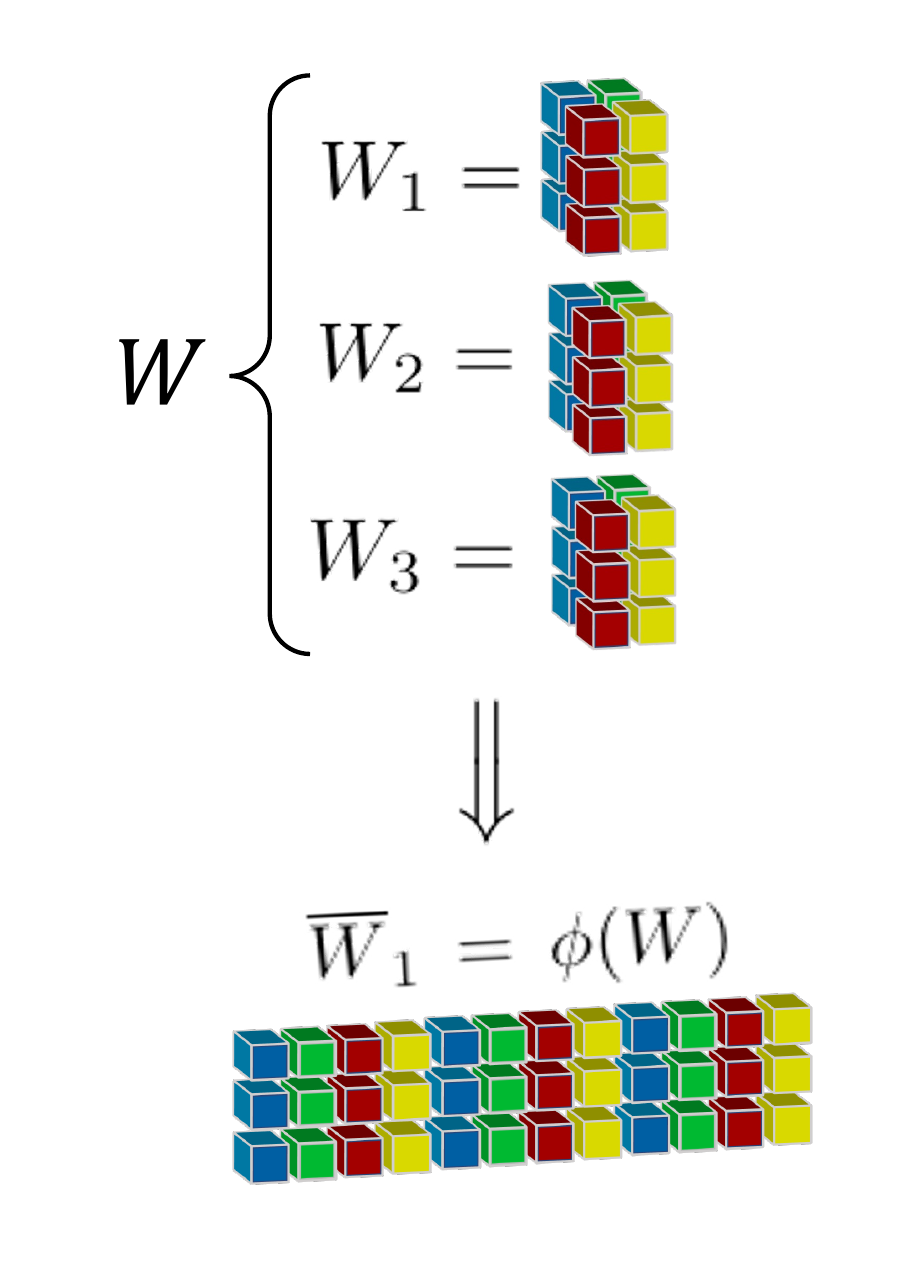}
	}
	\qquad
	\subfloat[$\wtwo$ is a reshaped multiple embedding of the weight tensor into a larger tensor space.]{
	    \label{fig:matrep2}
		\includegraphics[width=.6\linewidth]{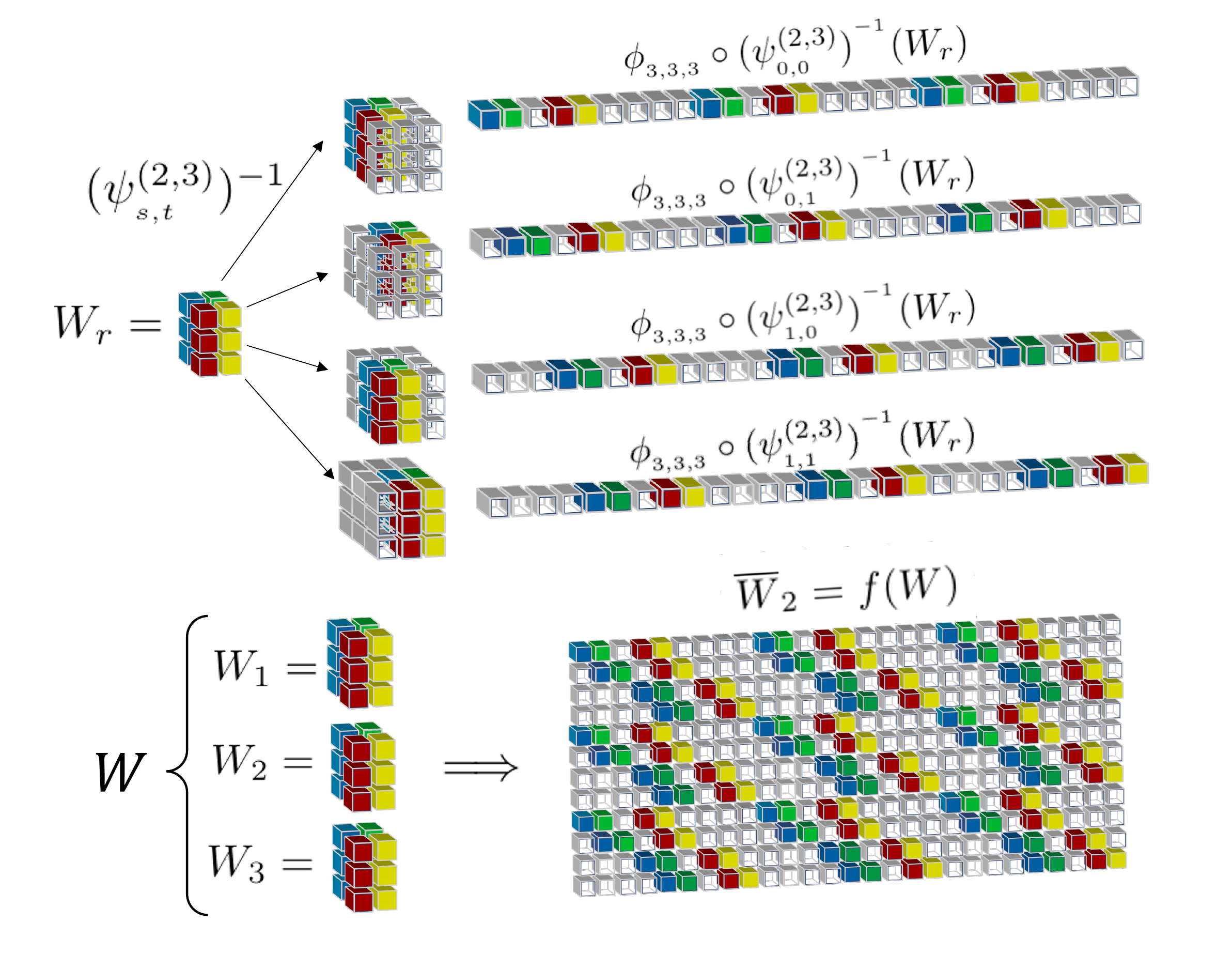}
	}
	\caption{\textbf{Illustration of the matrizations $\wone$ and $\wtwo$ of the weight tensor $W \in \T_{(3,3,2,2)}$.}  Uncolored cubes correspond to zeros and cubes of the same color correspond to elements of the same spatial position in each filter}
	\label{fig:matrep}
\end{figure}

\Cref{equation:yewx3} performs the dot product in $\R^{cm^2}$.
We flatten $X \in \T_{(c,m,m)}$ to $\phi(X) \in \R^{cm^2}$ and reshape $Y \in \T_{(d,n,n)}$ to $\ytwo = \phi(Y) \in \R^{dn^2}$.
Define an embedding
\begin{equation}
    f: \T_{(d,c,k,k)} \rightarrow \T_{(dn^2,cm^2)}
\end{equation}
and let $\wtwo = f(W)$ so that
if $\phi_{(d,n,n)}(Y)_r = Y_{h,i,j}$
 then the $r^{th}$ row
of $\wtwo$  is a reshaped embedding of the $h^{th}$ filter in $W$ into $\R^{cm^2}$:
\begin{equation}
    f(W)[r,:] = \wtwo[r,:] = \phi_{(c,m,m)}\circ (\psi_{i,j}^{(m,k)})^{-1}(W_{h}).
\end{equation}
This gives us a second matrix representation for cross-correlation  given by
\begin{equation}\label{equ:m2rep}
    \wtwo \cdot\phi(X) = \ytwo.
\end{equation}
Let $\bar{\chi}_2$ be the bilinear map defined by \cref{equ:m2rep} required to complete the commutative diagram in \cref{fig:cdiag2}.
Note that $\wtwo$ is a Toeplitz matrix representation of $W$. We illustrate the difference in the matrization of the tensors in \cref{fig:matrep}.

\Cref{equ:m1rep,equ:m2rep} produce the same output tensors up to a simple reshaping isomorphism.
Their difference lies in the coordinate systems in which their domains are represented, and in how we view their output.
The matrix $\wtwo$ can have up to $cm^2$ singular values. This is the matrization used in Sedghi, Gupta, and Long \cite{sedghi2019singular}.
The matrix $\wone$ can have up to $ck^2$ singular values. This is the map used by Yoshida and Miyato \cite{yoshida2017spectral}.\footnote{An interesting discussion of the differences between the perspectives can be found in \cite{sedghi2019singular}.}
Cross-correlation computes the spatial activations of $Y$ independently, sharing the same dynamics with respect to the singular vectors.
We claim that \cref{equ:m2rep} obfuscates these dynamics  by requiring the high dimensional representation,
 while \cref{equ:m1rep} preserves the spirit of shared weights and the translation invariance of convolution by placing the redundancies in $\Psi(X)$.

%% file: svdofw.tex
The unnormalized Gram matrix of the rows of $\wone$ is defined by the $d \times d$ matrix $\wone\wone^T$.
The entries in the Gram matrix $\wone\wone^T$ measure correlations between the linear maps defined by the rows in $\wone$, i.e., the filters in $W$.
This is analogous to the fully connected case.

In contrast, each filter is represented by $n^2$ rows in $\wtwo$ so that its Gram matrix has dimensions $dn^2 \times dn^2$.
Reordering the rows of $\wtwo$ to group the filters by their embeddings shows the Gram matrix $\wtwo\wtwo^T$ to have $n^2$ copies of $\wone\wone^T$ down the diagonal while the off-diagonal elements correspond to correlations between incommensurate embeddings of the filters. 
It is unclear what meaning these off-diagonal elements convey about the action of the weight tensor on $X$.

To better understand the difference between the two matrix representations we consider the distribution of their singular values.

\subsection{Validating the Matrix Representation of the Convolutional Layers}
\label{subsection:svdvalidation}
\input{sub_svdvalidation}

\subsection{Decomposition of a Simple CNN}
\label{subsection:decompmnist}
\input{mnist}

\subsection{The SVD of Weight Tensors}
\label{subsection:svdconv}
\input{tensorsvd}

%% file: sub_svdvalidation.tex
Martin and Mahoney \cite{martin2020heavytailed,martin2021implicit,martin2021predicting} apply random matrix theory to analyze the distribution of the singular values of the weight matrices of a typical model $\mcM$.
They demonstrate that well-trained generalizable models exhibit implicit self-regularization and these properties can be used to compute a metric for predicting test accuracy of CNNs without knowing the test data.
They note that modern models learn the correlations in the data and these correlations are stored in the weight matrices.
The empirical spectral distribution (ESD) of the Gram matrix of each linear layer $\Lm$ tends to exhibit a heavy-tailed power law fit.
The associated power law exponent $\alpha_{\Lm}$ provides a complexity metric for the layer $\Lm$; the smaller the value of $\alpha_{\Lm}$ the greater the regularization \cite{martin2021implicit}.
Using a weighted average of these exponents, the authors  define a capacity metric $\hat{\alpha}$ which is predictive of the test performance of the neural network \cite{martin2020heavytailed}:
\begin{equation}\label{eq:alphametric}
    \hat{\alpha}= \sum_{\Lm \in \mathscr{L}} \alpha_{\Lm} \log \lambda_{\Lm}^{\max },
\end{equation}
where $\lambda_{\Lm}^{\max}$ is the maximum eigenvalue for the Gram matrix for layer $\Lm$.
The smaller the value of  $\hat{\alpha}$ the better the test performance.

It is notable that the experimental results performed in \cite{martin2021implicit} are representative of the capacity metric restricted to fully connected layers.
This is in part because the choice of matrix representation for the convolutional layers was in question \cite{martin2021predicting}.
While the authors suggest the theory could be extended to convolutional layers and use it to compute $\hat{\alpha}$ for a large number of architectures, 
they do not fully extend their theoretical results to these layers. 
In particular, the matrices used to compute the capacity metric $\hat{\alpha}$ in \cite{martin2021predicting} 
were the channels of the weight kernels in the layer. Given the small size of the individual weight channels, it is unlikely that accurate power-law fits were obtained. Moreover, there is no a priori reason to believe that the correlations within a channel of the weight kernel is related to network capacity.

\begin{figure*}[t]
	\centering
	\subfloat[Singular values for the first layer of VGG-16 for $\wone \wone^T$ and $\wtwo \wtwo^T$. Note that the Toeplitz matrization $\wtwo$ does not have a heavy tail.]{
			\raisebox{-.6\height}{
			\includegraphics[width=.4\linewidth]{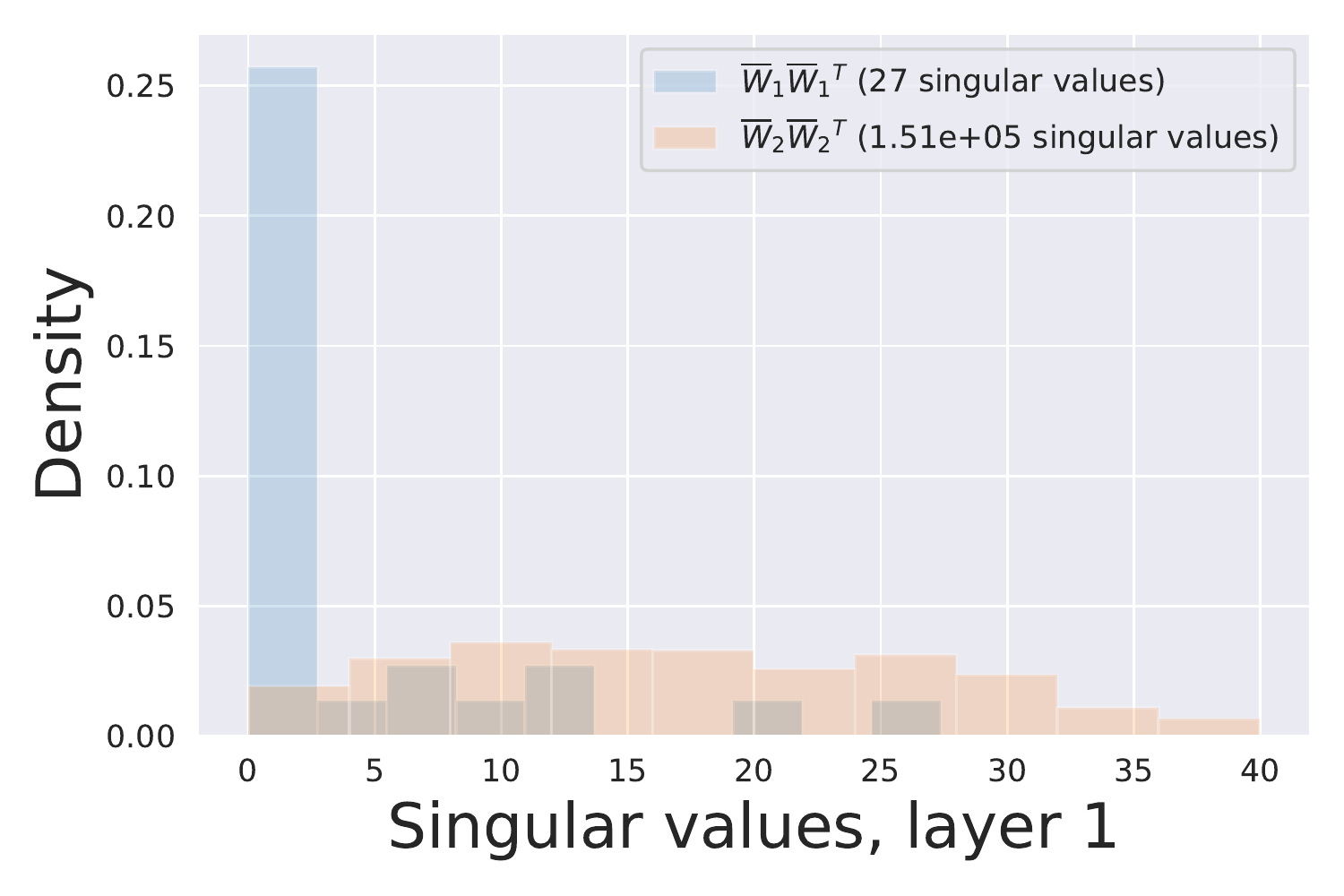}
			}
	}
	\qquad
	\captionsetup[subfloat]{captionskip=1.5em}
	\subfloat[The number of singular values and the layer $\alpha_{\Lm}$ metric for $\wone \wone^T$ and $\wtwo \wtwo^T$. The Toeplitz matrization $\wtwo$ has on the order of a million singular values for each layer.]{
	    \ra{1.05}
	    \begin{tabular}{rrrcrr@{}}
	        \toprule
            layer & \multicolumn{2}{c}{number of s-vals} && \multicolumn{2}{c}{$\alpha_{\Lm}$ metric} \\ 
            \cmidrule{2-3} \cmidrule{5-6} 
            & $\wone$ & $\wtwo$  && $\wone$  & $\wtwo$ \\ \hline
            2 & 64 & 3.21e6 && 1.66 & 4.63 \\
            3 & 128 & 3.21e6 && 1.8 & 3.77 \\
            4 & 128 & 6.42e6 && 1.87 & 4.48 \\
            5 & 256 & 1.61e6 && 3.9 & 5.05 \\
            6 & 256 & 3.21e6 && 2.07 & 3.42 \\
            7 & 256 & 3.21e6 && 4.42 & 2.92 \\
            8 & 512 & 3.21e6 && 5.58 & 3.27 \\
            9 & 512 & 6.42e6 && 4.14 & 2.57 \\
            10 & 512 & 1.61e6 && 3.03 & 2.28 \\
            11 & 512 & 1.61e6 && 4.81 & 2.58 \\
            12 & 512 & 1.61e6 && 4.09 & 2.48 \\
            13 & 512 & 1.61e6 && 4.02 & 1.75 \\
            \bottomrule
        \end{tabular}}
	\caption{\textbf{Validating our choice of matrix representation via the SVD for VGG-16 \cite{simonyan2015very} trained on ImageNet \cite{deng2009imagenet} -- Comparison of ESDs.} We compare the ESD of the Gram matrices for the two matrizations $\wone$ and $\wtwo$ across the convolutional layers. Note the large number of singular values for the centered Toeplitz matrization $\wtwo$, which has an unclear interpretation.}
	\label{fig:spectrum}
\end{figure*}

\begin{figure*}[t]
    \centering
    \subfloat[The $\hat{\alpha}$ computed with $\wone$ performs better than the original matrix choices used in \cite{martin2020heavytailed}.]{
        \label{subfig:testacc_w1}
		\includegraphics[width=.45\linewidth]{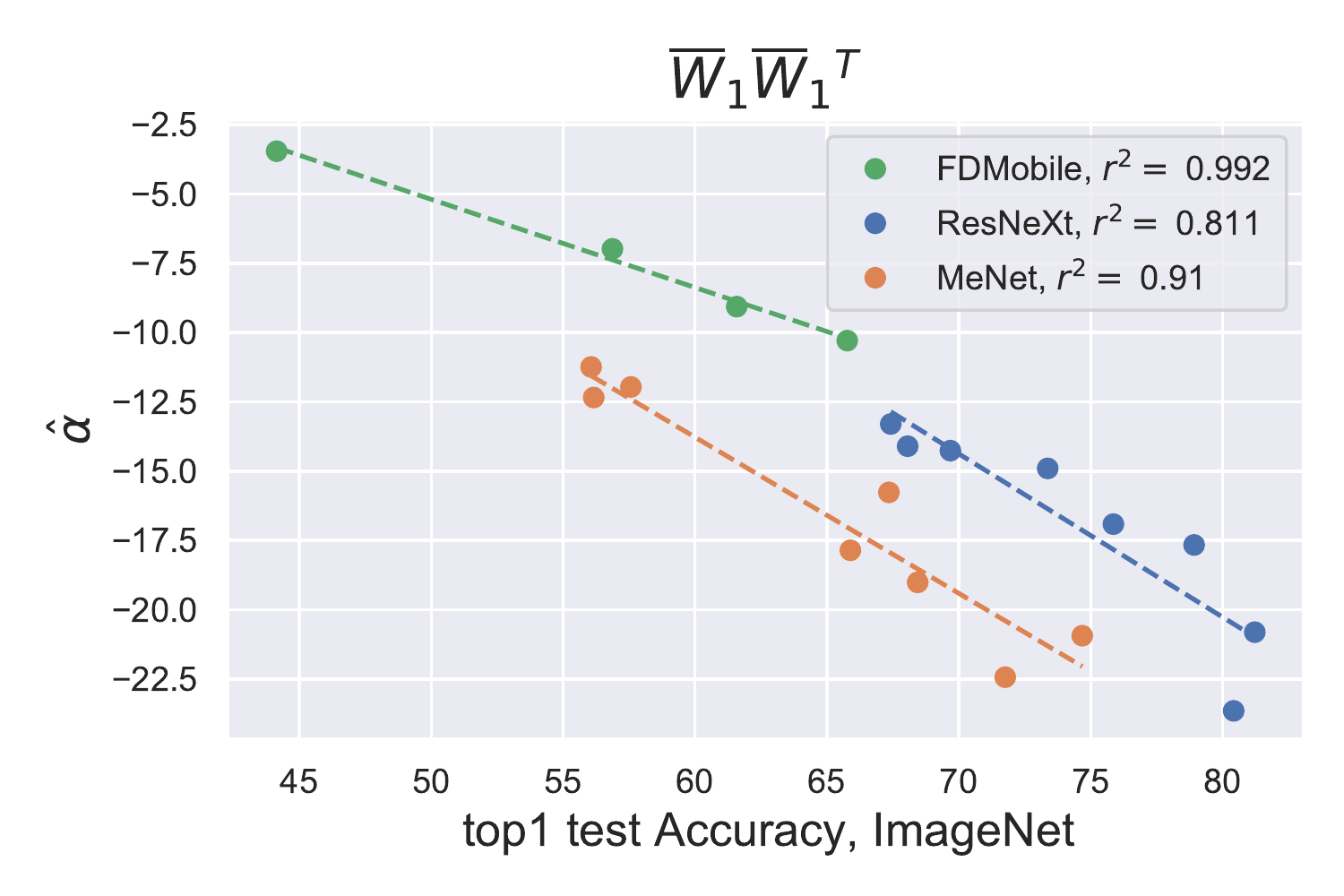}
	}
	\qquad
    \subfloat[The $\hat{\alpha}$ computed with the Toeplitz matrization $\wtwo$ are considerably less predictive of test accuracy.]{
        \label{subfig:testacc_w2}
		\includegraphics[width=.45\linewidth]{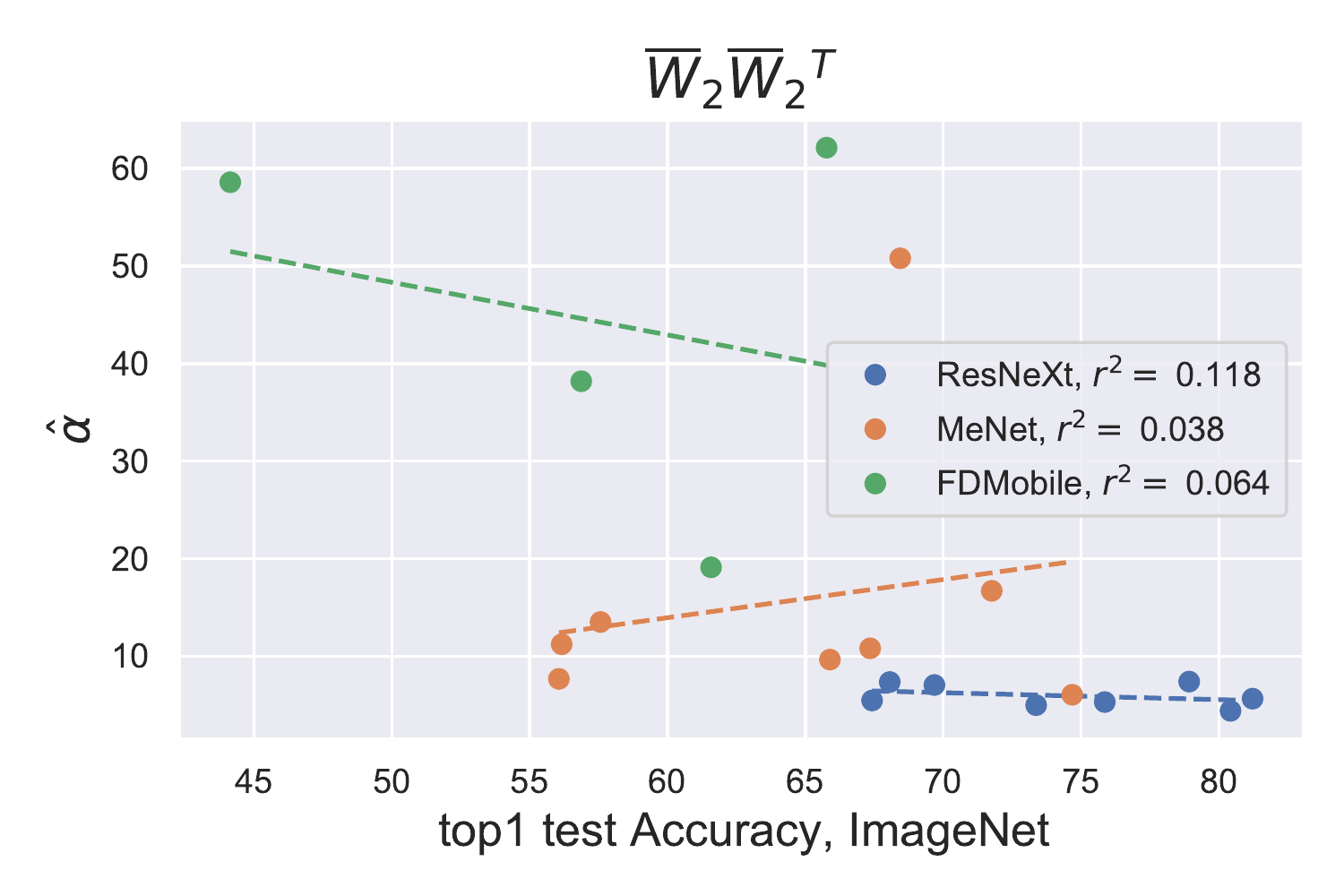}
    }
	\caption{\textbf{Validating our choice of matrix representation via the SVD for various models \cite{xie2017aggregated,qin2018fdmobilenet,xiong2019mixed} trained on ImageNet \cite{deng2009imagenet} -- Comparison of $\mathbf{\hat{\alpha}}$.} We compute the $\hat{\alpha}$ metric \cref{eq:alphametric} vs test accuracy for $\wone$ and $\wtwo$, and perform linear fits for each of the architecture classes and matrizations.}
	\label{fig:alphametric}
\end{figure*}

To validate our choice of matrix representation, we calculate $\hat{\alpha}$ for three different matrizations and across three classes of architectures (ResNeXt \cite{xie2017aggregated}, FD-MobileNet \cite{qin2018fdmobilenet}, and MeNet \cite{xiong2019mixed} models of varying depths) where the $\hat{\alpha}$ computed in \cite{martin2021predicting} yielded mixed results.

\Cref{fig:spectrum,fig:alphametric} show examples of the empirical spectral distributions and capacity metrics for models trained on ImageNet \cite{deng2009imagenet}.
\Cref{fig:spectrum} shows the spectral distribution for the Gram matrices associated with $\wone$ and $\wtwo$ for the convolutional layers of VGG-16 \cite{simonyan2015very}. For the first layer of VGG-16, the Gram matrix for $\wone$ has a spectral distribution consistent with the distribution found in the fully connected layers for the models tested, while $\wtwo$ does not.
In \cref{fig:alphametric}, we find that $\wone$ better aligns with theory and is considerably more predictive of test accuracy than $\wtwo$.                         
If we view the heavy tail as corresponding to extracted correlations from the data, then the features themselves should correlate to the corresponding singular vectors.
It is with this intuition that we use $\barw = \wone$ to represent $W$ as a matrix. Similarly, let $\bary = \yone$.

%% file: mnist.tex
CNN interpretability literature tends to look to the channel and spatial activations of hidden layer representations to explain the feature maps learned by the model
\cite{olah2018building,hohman2020summit,kim2018interpretability}.
Since the channel and spatial activations of $Y$  are projections onto a subset of the Euclidean basis for $\T_{(d,n,n)}$,
this is consistent with the assertion in \cite{bau2017network}  that the projection of activations onto a basis close to the Euclidean basis provides more meaningful representations of stored features than projections onto a random orthonormal basis.
But we have just observed there is a basis for each subspace of spatial activations derived from the singular vectors that holds the extracted features in terms of their singular vectors.
The relationship between the channel activations and the singular vectors for a layer $\Lm$ is described by the equation:
\begin{equation}\label{equation:channelsingular}
    \bary = \barw \cdot \Psi(X) = U S V^T \Psi(X).
\end{equation}
The factor $S V^T \psi(X) $ in the term on the right is a $d \times n^2$ matrix
such that the  $i^{th}$ row is the vector of correlations gotten from the receptive fields of $X$ with the $i^{th}$ right singular vector.
Each channel in $Y$ is a reshaped row of $\bary$, which is a linear combination of these vectors of correlations.

To illustrate this relationship and other concepts in this work we use
a simple CNN $\mathcal{N}$ for classifying the MNIST dataset of handwritten digits \cite{lecun2010mnist}.
The model $\mathcal{N}$ consists of four linear layers: two convolutional layers, $  conv1 $ and $ conv2 $, and two fully connected layers, $ fc1 $ and $ fc2 $.\footnote{The full architecture is given by the sequence: $conv1$, ReLU, $conv2$, ReLU, Maxpool, $fc1$, ReLU, $fc2$, Softmax.}

 \begin{figure*}[t] 		
	\centering
	\includegraphics[width=\linewidth]{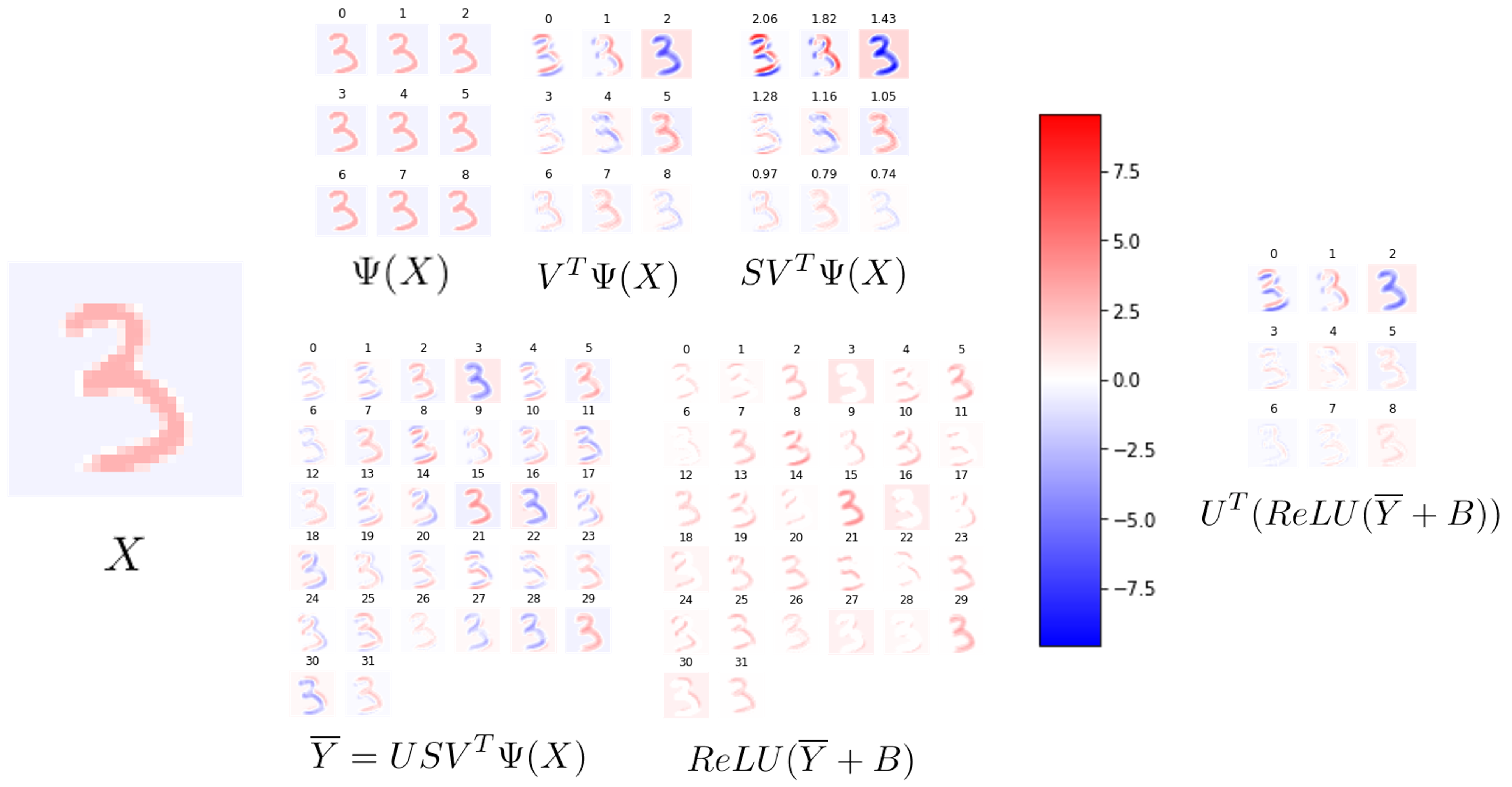}
	\caption{\textbf{Steps of computation for $\mathbf{conv1}$:} Each array of images represents a $2$-tensor or matrix.  Each small image is a heatmap of a reshaped row of the matrix, scaled for consistent display with a blue-red color scheme to best discern contrast. The input to the layer is the tensor $X$ and the output after ReLU is $ReLU(\bary+B)$. The singular values are shown with the corresponding projection above each image in $SV^T\Psi(X)$. To emphasize the dampening effect of ReLU on signal we also show the projection back onto the left singular vectors corresponding to nonzero singular values as $U^T(ReLU(\bary+B))$. The layer subscript $conv1$ is not shown on any of the variables.}
	\label{fig:mnist_layer1} 
\end{figure*}

In \cref{fig:mnist_layer1} we visualize each step of the decomposition for layer $conv1$.
Let $X \in \T_{(1,28,28)}$  represent a sample image from MNIST.
Layer $ conv1  $ has weight tensor $W_{conv1} \in \T_{32,1,3,3}$ and $\barw_{conv1} \in \T_{(32,9)}$.
Let $U_{conv1}S_{conv1}V_{conv1}^T$ be the SVD of $\barw_{conv1}$.
Each array of images represents a $2$-tensor or matrix reshaped for visualization purposes.
For example the receptive field matrix  $\Psi(X_{conv1}) \in \T_{(9,26^2)}$ has $9$ rows, each of length $676$, which are reshaped as $26 \times 26$ $2$-tensors and visualized as heatmaps rescaled for consistent display.
The matrix $V_{conv1}^T\Psi (X)$ shows the projection of the receptive fields on each of the nine singular vectors.
Matrix multiplication by $S_{conv1} \in \T_{(32,9)}$ rescales each row in $V_{conv1}^T\Psi(X)$ and embeds the $2$-tensor into $\T_{32,26^2}.$\footnote{We only show the projections onto singular vectors with nonzero singular values.}
Each heatmap is an indication of the strength of the signal in the direction of the corresponding singular vectors.
The matrix $\bary_{conv1} = U_{conv1}S_{conv1}V_{conv1}^T\Psi(X) $ shows the projection of $S_{conv1}V_{conv1}^T\Psi(X)$ onto the Euclidean basis.
Each small image of $\bary_{conv1}$ is the heatmap of a channel of $Y_{conv1} $.
The rows of $\bary_{conv1}$ are linear combinations of the rows in $S_{conv1}V_{conv1}^T\Psi(X)$.

In \cite{dittmer2020singular}, Dittmer, King, and Maass describe the relationship between the singular values pre- and post-ReLU in the linear layers of a multi-layer perceptron.
In particular they show that ReLU has the effect of dampening signal and that only inputs strongly correlated to the right singular vectors with the largest singular values will persist through ReLU.
We can extend their observation to the convolutional layers because the matrix multiplication of the fully connected layers is analogous to the matrix multiplication mapping receptive fields to spatial activations.
In \cref{fig:mnist_layer1}, $ReLU(Y_{conv1}+B_{conv1})$ is the tensor that passes out of the layer after ReLU is applied; it includes a small translation by a bias term $B_{conv1}$. 
To see the dampening effect ReLU had on the original signals we include $U^T(ReLU(Y_{conv1}+B_{conv1}))$, which is the projection of the layer's output onto the left singular vectors corresponding to nonzero singular values and is best compared with $S_{conv1}V^T_{conv1}\Psi(X)$.

%% file: tensorsvd.tex
With the above observations,  we put forth a simple definition for the SVD for the weight tensor used in convolutional layers.
Let $USV^T$ be an SVD of $\barw$.
We use the reshaping isomorphisms to transform each matrix in the decomposition into a $4$-tensor and replace matrix multiplication with tensor cross-correlation.
Define the reshaping maps as $\phi_{V^T} : \T_{(ck^2,ck^2)} \rightarrow \T_{(ck^2,c,k,k)}$,
 $\phi_S : \T_{(d,ck^2)} \rightarrow \T_{(d,ck^2,1,1)}$,
and $\phi_U : \T_{(d,d)} \rightarrow \T_{(d,d,1,1)}$.
Using a stride of 1 for the $\star$ operation, it is easily checked that for each $X \in \msD$
\begin{equation}
    W \star X  =   \phi_U(U) \star \left(\phi_S(S) \star( \phi_{V^T}(V^T) \star X)\right).
\end{equation}
While we could use this \textit{tensor SVD} for the rest of the discussion, we will stick with the matrix representation as it is more intuitive and reference the SVD for $\barw$ using:
\begin{equation}\label{equation:dotconvolution}
    \barw \cdot \Psi(X) = USV^T \Psi(X) = \bary.
\end{equation}
The column vectors $\{v_i\}$ of $V$ form an orthonormal basis for $\R^{ck^2}$ and the column vectors $\{u_i\}$ of $U$ form an orthonormal basis for $\R^d$.
The diagonal matrix $S \in \T_{(d,ck^2)}$ is non-negative and
its diagonal entries are ordered in descending order, $s_0 \geq s_{1} \geq  \dots \geq s_{r} \geq 0$, such that $\barw(v_i) = s_i u_i$ and $r = \text{min}\{d,ck^2\}$.

The matrices $U$ and $V$ need not be unique as there are potentially multiple bases that could be chosen, in particular when the $\{s_i\}$ contains duplicates, but the subspaces corresponding to each distinct singular value are unique.
Without loss of generality we assume there are no duplicate nonzero singular values so that each of the subspaces are $1$-dimensional copies of $\R$.\footnote{This assumption eases the notation without changing the arguments and has been observed to be true in practice.}

For each $v \in \R^{ck^2}$, there is a unique representation $v = \sum_{i=0}^{r-1} \langle v,v_i \rangle v_i$, where $\langle \cdot\rangle$ denotes the Euclidean inner product on $\R^{ck^2}$, and
\begin{equation}\label{equation:inner_product}
    \barw v = \sum_{i=0}^{r-1} \barw \langle v,v_i \rangle  v_i = \sum_{i=0}^{r-1} s_i \langle v,v_i \rangle u_i.
\end{equation}
The inner product $\langle v,v_i \rangle$ is a measure of correlation between the two vectors.
When $v$ is a receptive field (i.e. a column of $\Psi(X)$) we will call $s_i \langle v,v_i \rangle$ the \textit{signal} of $v$ in the direction of the singular vector $v_i$.

The sign of the signal indicates if the receptive field is positively or negatively correlated with $v_i$.
The value of $s_i$ indicates whether the correlation is increased or suppressed by the model before it is passed to a nonlinear activation and on to the next layer.
Since $\barw$ operates independently on each receptive field, the latent representations generated by the model are essentially defined by these signals.
From this we infer the discriminative power of a CNN lies in its singular vectors and interpretability might be achieved by determining the features of the domain which have strong positive correlation with the singular vectors of each layer.\footnote{We considered both positive and negative correlation but found positive correlations were the most informative using the framework we outline here. However, further exploration of negative correlations could prove valuable.}

We overload our notation a bit and let $X$ be the tensor representation for an image in $\msD$.
Let $\Lm \in \msL$ and $\barw_{\Lm}$ be the weight tensor for the layer.
Let $X_{\Lm}$ be the latent representation for $X$ used as input to the layer $\Lm$ when passing $X$ through the model.
Let $r_{\Lm}$ be the number of nonzero singular values for $\barw_{\Lm}$.
Define the signal vector $\sigma_{\Lm}(X) \in \R^{r_{\Lm}}$ so that the $i^{th}$ element of the vector is the average signal of all of
the receptive fields in $\Psi(X_{\Lm})$ in the direction of $v_i$.
The signal vectors provide a summary of signal strength in the direction of each singular vector.
We acknowledge this is a coarse summary as it loses density information of how much signal is concentrated in one part of the image;
nevertheless, strong average signals can still provide differentiation as we will see.

%% file: semhh.tex
For each $X \in \msD$  a singular vector $v_i$ from layer $\Lm$ is \textit{significant} for  $X$ with respect to a threshold $q$  if the $i^{th}$ element of $\sigma_{\Lm}(X)$ is greater than $q$.
A singular vector $v_i$ from layer $\Lm$ is \textit{significant for a class}  $C^j$ with respect to a threshold $q$ and a percentile $p$ if $v_i$ is significant for at least $p\%$ of the elements in $C^j$ with respect to $q$.
To be representative the signal should be significant for a majority of the class, so $p > 50$. We used $p=75$ for most of our examples.
Our thresholds were chosen for each layer by taking a high percentage quantile from the full distribution of signals for a representative sampling of the latent representations for images in $\mathscr{D}$. 
The greater the threshold, the fewer singular vectors will be significant for each image. 
Singular vectors significant for multiple classes are highly correlated to features common to those classes.
Singular vectors significant for a single class are highly correlated to some feature that is more prevalent in that class than in other classes.
By studying the many-to-many relationships between classes and the significant vectors in each layer we begin to define the discriminatory features used by the model.

\subsection{Hypergraphs}

Hypergraphs are generalizations of graphs which model the many-to-many relationships within data.
Hypergraphs preserve the important mutual relationships that can be lost in ordinary graphs \cite{berge1989hypergraphs}.
A hypergraph $\mathcal{H} = (\mathcal{V},\mathcal{E})$ consists of a set of nodes $\mathcal{V}$ and a set of hyperedges $\mathcal{E}$ such that each $e \in \mathcal{E}$ is a subset of $\mathcal{V}$.
While graph edges correspond to  exactly two nodes, hyperedges correspond to any number of nodes so that hypergraphs are often thought of as set systems but with more structure \cite{aksoy2020hypernetwork}.
We model the relationships between the target classes and the singular vectors significant for each class using hypergraphs.

Fix a threshold $q$ and a percentage $p\%$.
For each layer $\Lm \in \mathscr{L}$ we construct a hypergraph $\hl$ with
hyperedges indexed by the target classes $ \{C^j\} $ and nodes indexed by the layer's singular vectors $\{v_i \}$. 
A node $v_i$ belongs to a hyperedge $C^j$ if the singular vector $v_i$ is defined as significant for $p\%$ of the class $C^j$ with respect to the threshold $q$.
We discard the index for any empty hyperedges or singular vectors not significant for any class. 
The resulting hypergraph describes differences between classes in terms of the singular vectors most highly correlated to images in these classess.

The size of the hypergraph depends on the parameters $q$ and $p$.
The higher the threshold and the percentage, the fewer nodes and hyperedges and the more specific certain singular vectors will be to fewer classes. 
By varying both parameters we are able to generate a collection of hypergraphs of differing sizes.

\subsection{Semantic Hierarchy of a CNN}
\label{subsection:shcnn}

\begin{figure*}[t]
 	\centering
    \includegraphics[width=.85\linewidth]{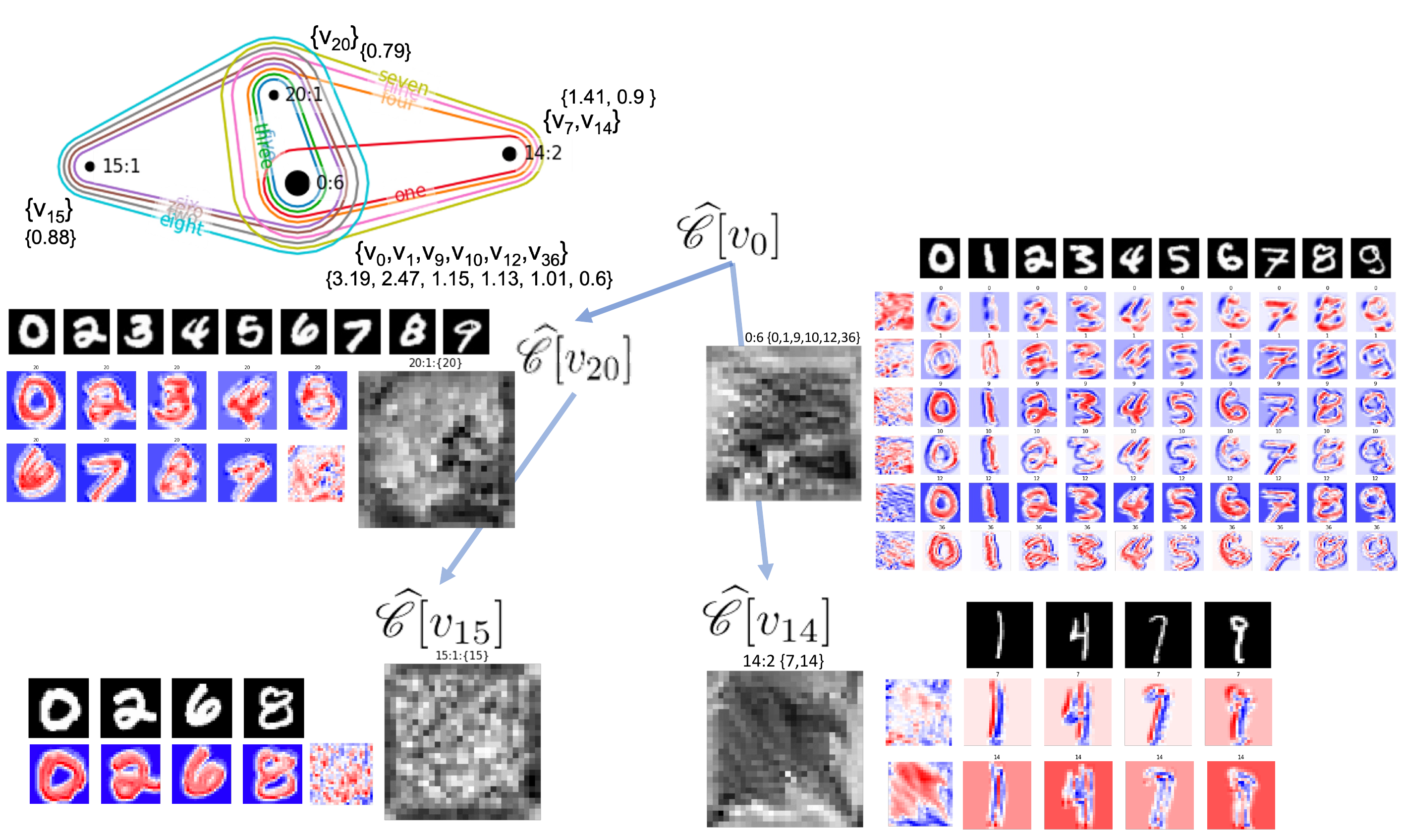}
    \caption{\textbf{Hypergraph and semantic hierarchy induced by $\mathbf{\hn{conv2}}$.} The hypergraph uses a $85\%$-quantile threshold and $75\%$ majority. Along with each optimized and exemplary image are its projections onto the subspaces spanned by the singular vectors in the equivalence class. Beside each equivalence class in the diagram are exemplary images from the corresponding target classes and a single image optimized using \cref{equation:fv1,equation:fv2}.}
    \label{subfig:conv2}
\end{figure*}

\begin{figure*}[t]
    \centering
    \includegraphics[width=.85\linewidth]{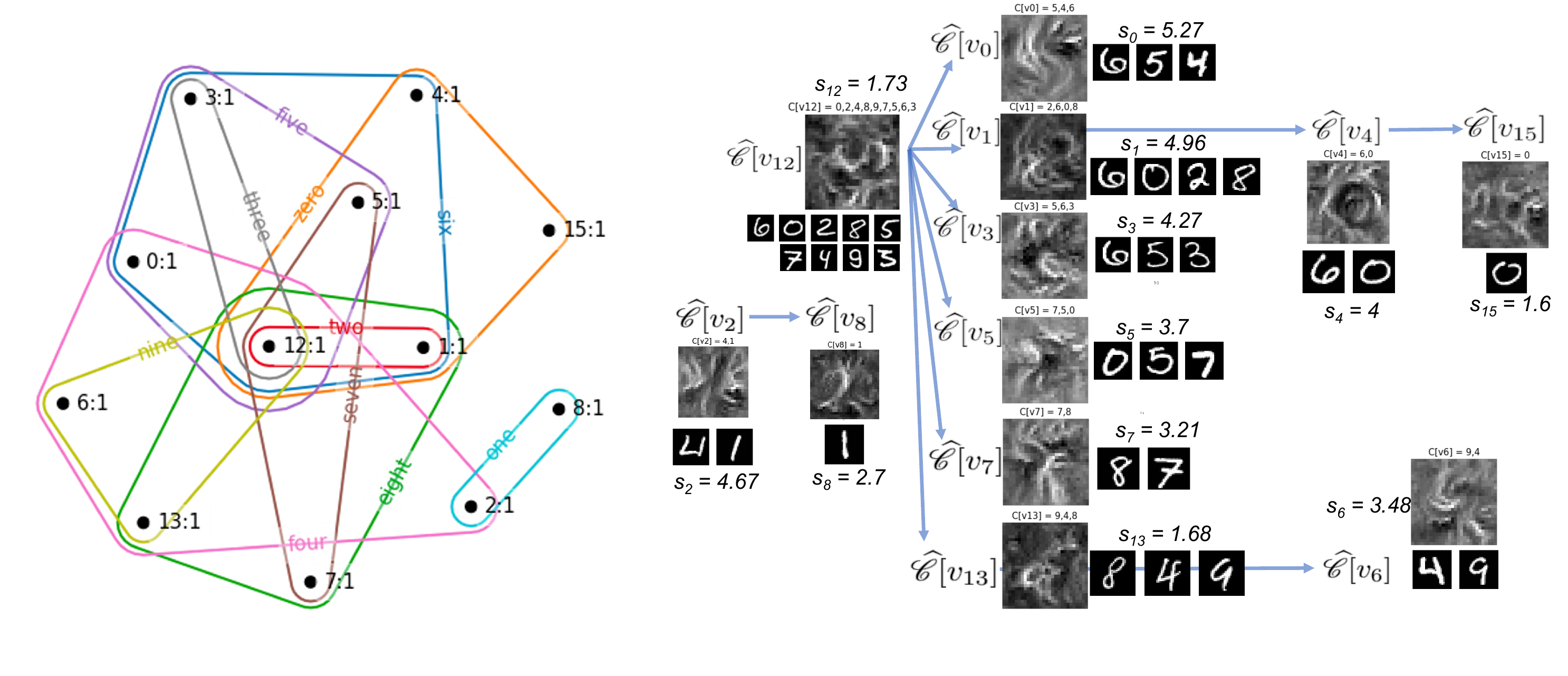}
    \caption{\textbf{Hypergraph and semantic hierarchy induced by $\mathbf{\hn{fc1}}$.} The hypergraph uses a a $95\%$-quantile threshold and $75\%$ majority. Beside each equivalence class in the diagram are exemplary images from the corresponding target classes and a single image optimized using \cref{equation:fv1,equation:fv2}.}
    \label{subfig:fc1}
 \end{figure*}
 
Let $\hl$ be the hypergraph of a linear layer $\Lm$ of model $\mcM$ defined by the parameters $q$ and $p$.
For each node $v_i$ in $\hl$, let $\widehat{\msC}[v_i]$ be the set of hyperedges to which $v_i$ belongs, i.e., the subset of target classes for which $v_i$ is significant.
Define an equivalence relation $\sim$ on the set of nodes in $\hl$ so that any two nodes, $v_i$, $v_j$, are equivalent 
if and only if they belong to the same set of hyperedges.
Let $[v_i]$ reference the equivalence class containing $v_i$.
Then
\begin{equation}
	v_i \sim v_j \iff \widehat{\msC}[v_i] = \widehat{\msC}[v_j] \iff [v_i] = [v_j].
\end{equation}

The hypergraph $\hl$ defines a partial ordering on the equivalence classes given by:
\begin{equation}
    [v_i]>[v_j] \iff \widehat{\msC}[v_i] \supset \widehat{\msC}[v_j].
\end{equation}
This partial ordering induces a \textit{semantic hierarchy} based on the features positively correlating to the singular vectors in each equivalence class.
Let $[v_i] \neq [v_j]$ be two equivalence classes. 
The concepts in the domain described by features positively correlated with the singular vectors in $[v_i]$ will be considered more general than concepts from features positively correlated with the singular vectors in $v_j$ if $[v_i] > [v_j]$.
Diagrammatically we represent this as $\widehat{\msC}[v_i] \longrightarrow \widehat{\msC}[v_j]$. We use this notation to diagram the semantic hierarchy induced by the hypergraph as in \cref{subfig:conv2,subfig:fc1,fig:cifar14}.

To illustrate, we construct two hypergraphs $\hn{conv2}$ and $\hn{fc1}$ for the model
$\mathcal{N}$ trained on MNIST of \cref{subsection:decompmnist}. 
We use a test set with $50$ images in each class using thresholds corresponding to the top $85\%$- and $95\%$-quantiles respectively for at least $75\%$ of the classes. 
We visualize the hypergraphs in \cref{subfig:conv2,subfig:fc1}.\footnote{Hypergraph figures were generated by the HyperNetX library \cite{hypernetx}.}
Rather than displaying all of the nodes in the hypergraph, the nodes are grouped by equivalence class and labeled by the index of one singular vector in the class followed by the number of nodes in the class.
Next to each hypergraph we show the corresponding diagram depicting the semantic hierarchy induced by the hypergraph along with exemplary images from each class.
We use these diagrams to direct our interpretability analysis to the singular vectors most significant for each class and hence most influential for discriminating the class.

%% file: framework.tex
One goal of model interpretability is to identify the domain-centric features the model uses for classification.
We have seen that any enhancement or suppression of a feature from the data depends on the correlation of its latent representations with the singular vectors.
From the point of view of the model, the features used for classification are the ones with latent representations most highly correlated to the singular vectors with the greatest singular values.

In this section we suggest a methodology for interpreting image classification CNNs using the framework defined by the hypergraphs and semantic hierarchies of the model described in \cref{sec:semhh,subfig:conv2,subfig:fc1,fig:cifar14}.
The diagram generated by the semantic hierarchy indicates how the model responds to images of different classes.
Features corresponding to singular vectors particularly significant for one class but not for another characterize how the model differentiates classes.

Using existing visualization techniques, we look for features highly correlated with the singular vectors and their equivalence classes.
By identifying the domain-centric features correlated to each equivalence class we identify the features that provide linear separability within the model.

\begin{figure*}[ht]
	\centering
	\includegraphics[width=.8\linewidth]{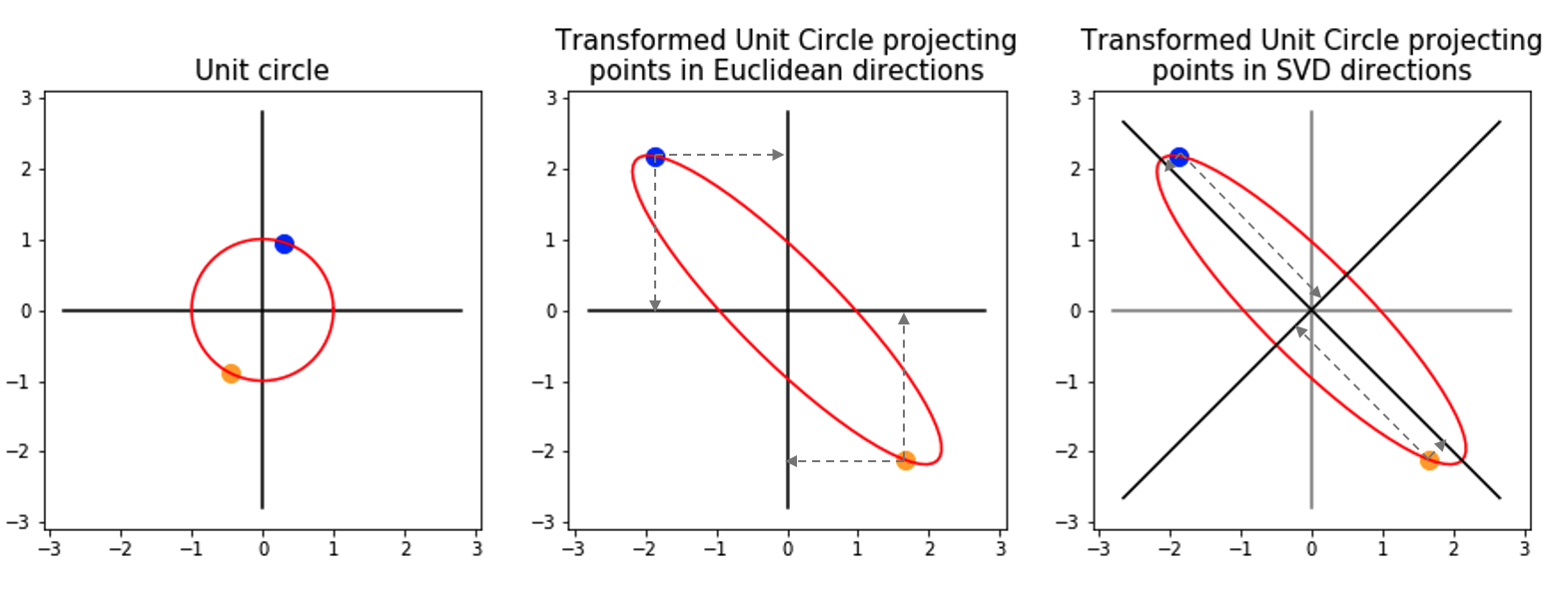}
	\caption{Projections of linearly transformed circle show that the dynamics (shrinking and stretching) of the map are better captured when projecting in the direction of the singular vectors.}
	\label{fig:circtrans}
\end{figure*}

This is very much in the spirit of Olah et al. \cite{olah2017feature,olah2018building} where features are discovered by optimizing images with latent representations positively correlated to the channels of the weight tensors.
The difference between projecting on the Euclidean directions and the singular vectors is illustrated in  \cref{fig:circtrans}.
Projection onto the singular vectors preserves the dynamical changes imposed by the layer on its input.
Much of this information can be lost when projecting onto the Euclidean basis vectors.
By optimizing for a specific singular vector we look for features in the domain that the model either deems important, hence increases their value in the latent representation, or unimportant, hence decreases their value.
This provides meaning to the learned parameters and explains the decision process in terms of the dynamics of the linear layers.

We generate an optimal image $X$ with a latent representation $X_{\Lm}$, which highly correlates with a particular singular vector $v_i$ in layer $\Lm \in \msL$ by computing:
\begin{equation}\label{equation:fv1}
    \begin{split}
        X_{opt} 	&=	\argmax_{T \in \T_{(c,m,m)}} \sum_j (s_i v_i ^T \Psi(T_{\Lm}))[j]\\
	    &=	\argmax_{T \in \T_{(c,m,m)}} \sum_j (u_i^T \barw\Psi(T_{\Lm}))[j]
    \end{split}
\end{equation}
subject to the transformation robustness constraints from \cite{olah2017feature}.

We generate an image maximally triggering a collection of singular vectors $\{v_i\}$ using a sum of weighted terms:
\begin{equation}\label{equation:fv2}
    \begin{split}
        X_{opt} &=\argmax_{T \in \T_{(c,m,m)}} \sum_{i}  w_{i} \left(  \sum_j (s_i v_i ^T \Psi(T_{\Lm}))[j]\right)\\
        &=\argmax_{T \in \T_{(c,m,m)}} \sum_{i}  w_{i}\left( \sum_j (u_i^T \barw\Psi(T_{\Lm}))[j]\right)
    \end{split}
\end{equation}
for some set of weights $\{w_{i}\}$.

As tantalizing as these images can sometimes be, we must be careful.
If the objective function simply adds the signals across the spatial activations as we did in \cref{equation:fv1,equation:fv2} it could smear signal out across an input image and lose the density of the signal actually found in the images.
Recognizing the limitations of feature visualization,
we supplement our analysis by identifying exemplary images, which highly activate the signal or group of signals in an equivalence class.
This is done by restricting \cref{equation:fv1,equation:fv2} to tensors in $\T_{(c,m,m)}$ representing images in the model's test or training set.
In early layers where latent representations retain much of the spatial information of the original image, we examine the scaled projects of spatial activations from the latent representations onto the singular vectors and study the heatmap of signals produced.
The heatmaps define spatial regions with features highly correlated with the singular vectors similar to a saliency map.
By comparing optimized images and heatmaps of projections with exemplary images we hope to gain intuition around the domain-centric features deemed important by the model.
Where a dataset has large enough image resolution, similarity overlays from \cite{fong2021interactive} can also be used to relate the feature visualizations and exemplary images.

\subsection{Interpreting \texorpdfstring{{$\mathcal{N}$}}{N} with MNIST}
\label{subsection:intmnist}
\input{mnist_int}

\subsection{DeepDataProfiler}
\label{subsection:ddp}

\begin{figure*}[t]
	\centering
	\includegraphics[width=.9\linewidth]{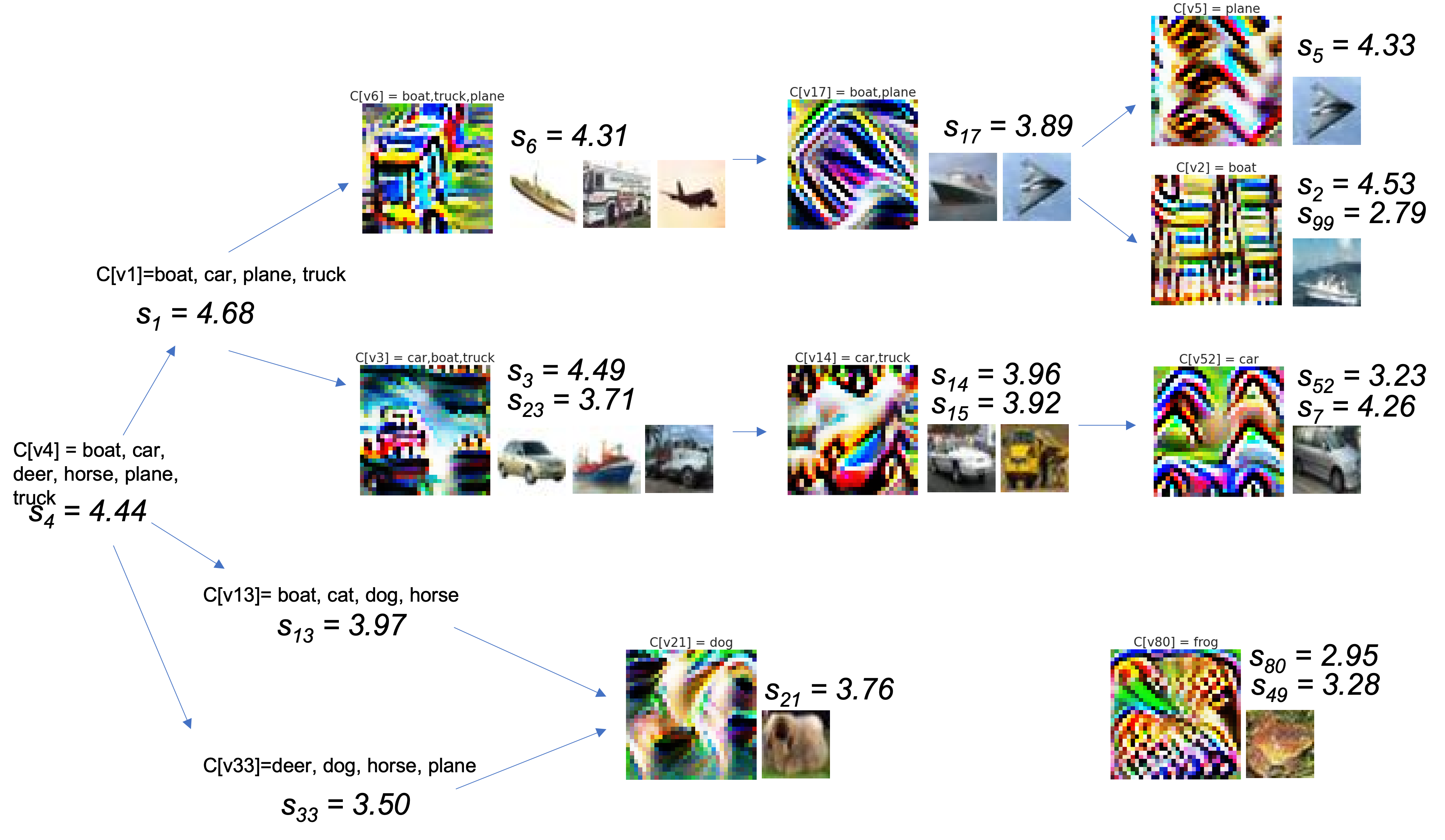}
	\caption{\textbf{Exploration of VGG-16 \cite{simonyan2015very} on CIFAR-10 \cite{lecun2010mnist} -- Diagram for part of the semantic hierarchy induced by a hypergraph for the 7\textsuperscript{th} convolutional layer.} The hypergraph was created using a significance threshold equal to the $90\%$-quantile of all signals and percentage majority of greater than $50\%$ for each class.}
	\label{fig:cifar14}
\end{figure*}

\begin{figure*}[t]
    \centering
	\includegraphics[width= .9\linewidth]{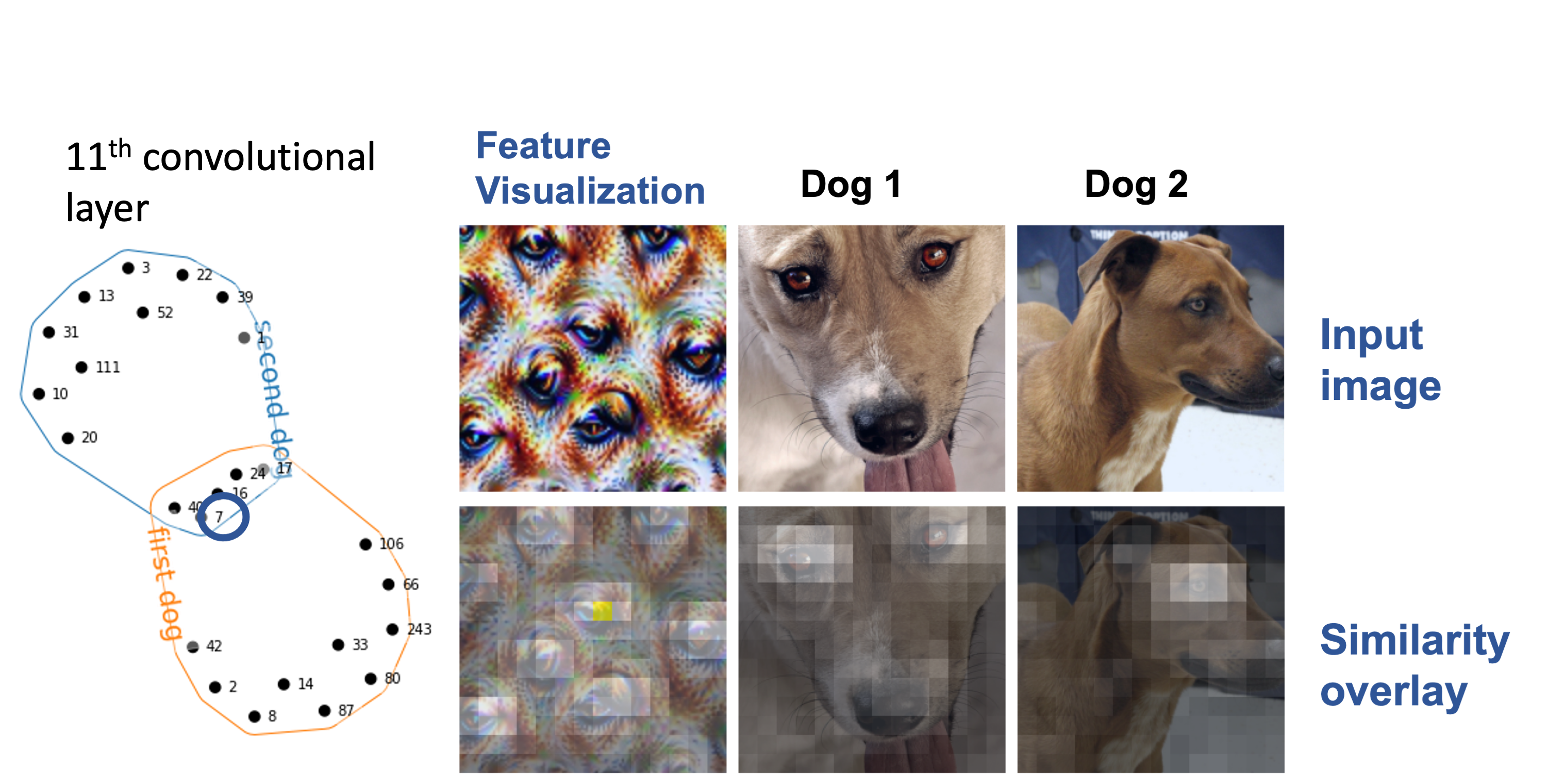}
	\caption{\textbf{Exploration of VGG-16 \cite{simonyan2015very} on ImageNet \cite{deng2009imagenet} -- Significant features shared by two images.}  The hypergraph models the  significant singular vectors for two dog images (using $50\%$-quantile) from the 11\textsuperscript{th} convolutional layer. Exemplary images and feature visualization are for the singular vector $v_7$. Similarity overlays highlight cosine similarity between the latent representations of the three images.}
	\label{fig:dogeyes}
\end{figure*}

We use this framework to guide our exploration of VGG-16 \cite{simonyan2015very} trained on CIFAR-10 \cite{krizhevsky2009learning} and ImageNet \cite{deng2009imagenet} in our DeepDataProfiler (DDP) library available on GitHub \cite{deepdataprofiler}.
DDP is a Python library for generating attribution graphs called \textit{profiles} for convolutional neural networks.
The library includes a module for studying the SVD of convolutional layers along with Jupyter notebooks \cite{kluyver2016jupyter} for generating hypergraphs and feature visualizations.

Our goal is to understand the features used by the network to differentiate images in each of the datasets.
The semantic hierarchy guides our inquiry by recognizing the features most strong for each image and class.
For example, \cref{fig:cifar14} shows part of the semantic hierarchy for the 7\textsuperscript{th} convolutional layer of VGG-16 when trained on CIFAR-10.
Each level of the diagram displays an optimized image for the corresponding equivalence class, exemplary images from CIFAR-10, and the values of the corresponding singular vectors.
One observable difference gleaned from the figure is the softer edges in the feature visualizations connected with living things versus the objects for transportation.

Feature visualizations for singular vectors associated with VGG-16 trained on ImageNet have superior resolution and tend to be more interpretable.
\Cref{fig:dogeyes} shows a feature visualization for a singular vector significant to two images from the dog class in ImageNet.
The similarity overlay highlights the cosine similarity of the spatial activations in the 11\textsuperscript{th} convolutional layer between the optimized image and the two dog images.
Exemplary images and feature visualizations for each layer of VGG-16 trained on ImageNet are available from our Streamlit application \cite{ddpvis}, which supports the DDP library.

%% file: mnist_int.tex
In \cref{subfig:conv2}, hypergraph $\hn{conv2} $ has 10 nodes divided into 4 equivalence classes. 
Equivalence class $[v_0]$ is at the top of the semantic hierarchy, because its elements are significant for every target class in MNIST \cite{lecun2010mnist} with respect to the $85\%$-quantile. 
This means the features, which are increased or suppressed, are common to images in all of the classes. 
We also show optimized and exemplary images for each of the equivalence classes in $\hn{conv2}$. 
We found feature visualization to be more informative in the later layers of larger networks, where the receptive fields correspond to greater regions in the original image.
For early layers and in shallow networks, much can be learned by visualing the scaled projections of the latent representations of exemplary images onto the significant singular vectors in each equivalence class. 
For each exemplary image with tensor representation $X$ and each significant vector $v_i$ we compute the dot product of $s_iv_i$ with each of the columns of $\Psi(X)$.
The resulting vectors are reshaped into $26 \times 26$ 2-tensors and visualized in red, white, and blue. 
The color scale is set to show strong positive correlation with dark red and strong negative correlation in dark blue as is seen in the diagram.

Of particular interest are the features exhibited for $[v_{14}] = \{v_7,v_{14}\}$ and $[v_{15}] = \{v_{15}\}$.
Both their exemplary images and their projections are quite different. 
The scaled projections onto $v_{14}$ positively correlate with large black regions in the image and are slightly suppressed by the singular value 0.9.
The scaled projections onto $v_7$ also positively correlate with large black regions but the signal is stronger where the black is just to the left of the white of the image.
These regions are darker red and were slightly enhanced by the singular value 1.41.
Images with high positive correlation to $v_{15}$ have more white in their receptive fields and narrow bands of black around the edges. Their signals were slightly suppressed by the singular value 0.88. 
This branch in the semantic hierarchy of the first layer appears to separate images in part based on what portion of the image is background.

With deeper layers more complex shapes arise in the feature visualizations. Even in such a simple model as  $\mcN$ we see more discernible shapes in the optimized images for $fc1$ than we did in $conv2$. In \cref{subfig:fc1} we see the greatest singular values, $s_0$ and $s_1$ are associated with recognizable forms including an \textit{S}-curve and figure eight.

%% file: conclusion.tex
In this work we examine the linear transformations of the convolutional layers of a CNN.
We describe a collection of isomorphisms and embeddings on the vector space of tensors over a fixed index set and define convolution as a bilinear map on these tensor spaces that can be represented as matrix multiplication.
Our approach, which maps a 4-dimensional weight tensor of the convolutional layers isomorphically to a 2-dimensional matrix, provides an alternative to the commonly used Toeplitz representation for treating convolution as matrix multiplication.
This simple unfolding of the weight tensor preserves the spirit of shared weights and exposes the dynamics of the linear map as it acts on the space of receptive fields.
 
Our choice of matrization is validated both notionally by looking at the corresponding Gram matrix and empirically by applying results from random matrix theory.
The singular vectors of the matrix representation for the weight tensors identify the features of the input domain that the model has learned to either increase or suppress depending on their associated singular values. 

We define the significance of the singular vectors to each of the target classes using two parameters and model this relationship using hypergraphs.
By varying the parameters we generate a family of hypergraphs, each inducing a semantic hierarchy.
We describe how these hierarchies might be used to explore the model and to discover the discriminative features it uses for classification.
We introduce the DeepDataProfiler library, which uses these matrizations to gain interpretability of CNNs. 

This work brings fresh eyes to interpretability research by applying the same mathematical tools to convolutional layers that have been used to explain linear transformations for more than a century, giving us an opportunity to remove the black box stigma that haunts CNN models.
By identifying the features the model deems important we can study their inter-relationships and determine their discriminatory value and relevance to the data domain.
The greatest challenge moving forward will be to refine our optimization techniques, which translate the model's representation of these features into something we can recognize.
Future work should also explore the range of singular vectors to better understand how the model uses negative correlation and identifies noise.

As a final note, the implication of this new perspective extends beyond the scope of this paper.
The SVD of convolutional layers offers us an opportunity to prune convolutional models and identify misclassifications analogous to the work of Dittmer, King, and Maass \cite{dittmer2020singular} on multi-layer perceptrons.

%% file: main.bbl
\begin{thebibliography}{10}

\bibitem{aksoy2020hypernetwork}
Sinan~G. Aksoy, Cliff~A. Joslyn, Carlos~Ortiz Marrero, Brenda Praggastis, and
  Emilie Purvine.
\newblock Hypernetwork science via high-order hypergraph walks.
\newblock {\em {EPJ} Data Sci.}, 9(1):16, 2020.

\bibitem{andrews1976svd}
Harry~C. Andrews and Claude L.~Patterson III.
\newblock Singular value decomposition {(SVD)} image coding.
\newblock {\em {IEEE} Trans. Commun.}, 24(4):425--432, 1976.

\bibitem{bau2017network}
David Bau, Bolei Zhou, Aditya Khosla, Aude Oliva, and Antonio Torralba.
\newblock Network dissection: Quantifying interpretability of deep visual
  representations.
\newblock In {\em 2017 {IEEE} Conference on Computer Vision and Pattern
  Recognition, {CVPR} 2017, Honolulu, HI, USA, July 21-26, 2017}, pages
  3319--3327. {IEEE} Computer Society, 2017.

\bibitem{berge1989hypergraphs}
Claude Berge.
\newblock {\em Hypergraphs - combinatorics of finite sets}, volume~45 of {\em
  North-Holland mathematical library}.
\newblock North-Holland, 1989.

\bibitem{ddpvis}
Davis Brown, Madelyn Shapiro, and Brenda Praggastis.
\newblock {PNNL DeepDataProfiler} visualization tool.
\newblock Streamlit web app, 2022.

\bibitem{brunton2022data}
Steven~L. Brunton and J.~Nathan Kutz.
\newblock {\em Data-Driven Science and Engineering: Machine Learning, Dynamical
  Systems, and Control}.
\newblock Cambridge University Press, 2 edition, 2022.

\bibitem{deng2009imagenet}
Jia Deng, Wei Dong, Richard Socher, Li{-}Jia Li, Kai Li, and Li~Fei{-}Fei.
\newblock Imagenet: {A} large-scale hierarchical image database.
\newblock In {\em 2009 {IEEE} Computer Society Conference on Computer Vision
  and Pattern Recognition {(CVPR} 2009), 20-25 June 2009, Miami, Florida,
  {USA}}, pages 248--255. {IEEE} Computer Society, 2009.

\bibitem{dey2022human}
Sanjoy Dey, Prithwish Chakraborty, Bum~Chul Kwon, Amit Dhurandhar, Mohamed~F.
  Ghalwash, Fernando J.~Suarez Saiz, Kenney Ng, Daby Sow, Kush~R. Varshney, and
  Pablo Meyer.
\newblock Human-centered explainability for life sciences, healthcare, and
  medical informatics.
\newblock {\em Patterns}, 3(5):100493, 2022.

\bibitem{dittmer2020singular}
S{\"{o}}ren Dittmer, Emily~J. King, and Peter Maass.
\newblock Singular values for {ReLU} layers.
\newblock {\em {IEEE} Trans. Neural Networks Learn. Syst.}, 31(9):3594--3605,
  2020.

\bibitem{fan2020interpretability}
Fenglei Fan, Jinjun Xiong, Mengzhou Li, and Ge~Wang.
\newblock On interpretability of artificial neural networks, 2020.

\bibitem{fong2021interactive}
Ruth Fong, Alexander Mordvintsev, Andrea Vedaldi, and Chris Olah.
\newblock Interactive similarity overlays.
\newblock In {\em VISxAI}, 2021.

\bibitem{hohman2020summit}
Fred Hohman, Haekyu Park, Caleb Robinson, and Duen Horng~(Polo) Chau.
\newblock Summit: Scaling deep learning interpretability by visualizing
  activation and attribution summarizations.
\newblock {\em {IEEE} Trans. Vis. Comput. Graph.}, 26(1):1096--1106, 2020.

\bibitem{kim2018interpretability}
Been Kim, Martin Wattenberg, Justin Gilmer, Carrie~J. Cai, James Wexler,
  Fernanda~B. Vi{\'{e}}gas, and Rory Sayres.
\newblock Interpretability beyond feature attribution: Quantitative testing
  with concept activation vectors {(TCAV)}.
\newblock In Jennifer~G. Dy and Andreas Krause, editors, {\em Proceedings of
  the 35th International Conference on Machine Learning, {ICML} 2018,
  Stockholmsm{\"{a}}ssan, Stockholm, Sweden, July 10-15, 2018}, volume~80 of
  {\em Proceedings of Machine Learning Research}, pages 2673--2682. {PMLR},
  2018.

\bibitem{kluyver2016jupyter}
Thomas Kluyver, Benjamin Ragan{-}Kelley, Fernando P{\'{e}}rez, Brian~E.
  Granger, Matthias Bussonnier, Jonathan Frederic, Kyle Kelley, Jessica~B.
  Hamrick, Jason Grout, Sylvain Corlay, Paul Ivanov, Dami{\'{a}}n Avila, Safia
  Abdalla, Carol Willing, and Jupyter~Development Team.
\newblock Jupyter notebooks - a publishing format for reproducible
  computational workflows.
\newblock In Fernando Loizides and Birgit Schmidt, editors, {\em Positioning
  and Power in Academic Publishing: Players, Agents and Agendas, 20th
  International Conference on Electronic Publishing, G{\"{o}}ttingen, Germany,
  June 7-9, 2016}, pages 87--90. {IOS} Press, 2016.

\bibitem{kolda2009tensor}
Tamara~G. Kolda and Brett~W. Bader.
\newblock Tensor decompositions and applications.
\newblock {\em {SIAM} Rev.}, 51(3):455--500, 2009.

\bibitem{krizhevsky2009learning}
Alex Krizhevsky.
\newblock Learning multiple layers of features from tiny images.
\newblock Technical report, 2009.

\bibitem{lecun2010mnist}
Yann LeCun, Corinna Cortes, and CJ~Burges.
\newblock {MNIST} handwritten digit database.
\newblock {\em ATT Labs [Online]}, 2, 2010.

\bibitem{mahendran2015understanding}
Aravindh Mahendran and Andrea Vedaldi.
\newblock Understanding deep image representations by inverting them.
\newblock In {\em {IEEE} Conference on Computer Vision and Pattern Recognition,
  {CVPR} 2015, Boston, MA, USA, June 7-12, 2015}, pages 5188--5196. {IEEE}
  Computer Society, 2015.

\bibitem{martin2020heavytailed}
Charles~H. Martin and Michael~W. Mahoney.
\newblock Heavy-tailed universality predicts trends in test accuracies for very
  large pre-trained deep neural networks.
\newblock In Carlotta Demeniconi and Nitesh~V. Chawla, editors, {\em
  Proceedings of the 2020 {SIAM} International Conference on Data Mining, {SDM}
  2020, Cincinnati, Ohio, USA, May 7-9, 2020}, pages 505--513. {SIAM}, 2020.

\bibitem{martin2021implicit}
Charles~H. Martin and Michael~W. Mahoney.
\newblock Implicit self-regularization in deep neural networks: Evidence from
  random matrix theory and implications for learning.
\newblock {\em Journal of Machine Learning Research}, 22(165):1--73, 2021.

\bibitem{martin2021predicting}
Charles~H. Martin, Tongsu~(Serena) Peng, and Michael~W. Mahoney.
\newblock Predicting trends in the quality of state-of-the-art neural networks
  without access to training or testing data.
\newblock {\em Nature Communications}, 12(4122), 2021.

\bibitem{nguyen2016multifaceted}
Anh~Mai Nguyen, Jason Yosinski, and Jeff Clune.
\newblock Multifaceted feature visualization: Uncovering the different types of
  features learned by each neuron in deep neural networks, 2016.

\bibitem{olah2020overview}
Chris Olah, Nick Cammarata, Ludwig Schubert, Gabriel Goh, Michael Petrov, and
  Shan Carter.
\newblock An overview of early vision in {InceptionV1}.
\newblock {\em Distill}, 2020.

\bibitem{olah2020zoom}
Chris Olah, Nick Cammarata, Ludwig Schubert, Gabriel Goh, Michael Petrov, and
  Shan Carter.
\newblock Zoom in: An introduction to circuits.
\newblock {\em Distill}, 2020.

\bibitem{olah2017feature}
Chris Olah, Alexander Mordvintsev, and Ludwig Schubert.
\newblock Feature visualization.
\newblock {\em Distill}, 2017.

\bibitem{olah2018building}
Chris Olah, Arvind Satyanarayan, Ian Johnson, Shan Carter, Ludwig Schubert,
  Katherine Ye, and Alexander Mordvintsev.
\newblock The building blocks of interpretability.
\newblock {\em Distill}, 2018.

\bibitem{hypernetx}
Brenda Praggastis, Dustin Arendt, Cliff Joslyn, Emilie Purvine, Sinan Aksoy,
  and K~Monson.
\newblock {HyperNetX}, 2022.
\newblock Version 1.2.4.

\bibitem{deepdataprofiler}
Brenda Praggastis, Davis Brown, and Madelyn Shapiro.
\newblock {PNNL DeepDataProfiler}, 2022.
\newblock Version 2.0.0.

\bibitem{psichogios1994svdnet}
Dimitris~C. Psichogios and Lyle~H. Ungar.
\newblock {SVD-NET:} an algorithm that automatically selects network structure.
\newblock {\em {IEEE} Trans. Neural Networks}, 5(3):513--515, 1994.

\bibitem{qin2018fdmobilenet}
Zheng Qin, Zhaoning Zhang, Xiaotao Chen, Changjian Wang, and Yuxing Peng.
\newblock Fd-mobilenet: Improved mobilenet with a fast downsampling strategy.
\newblock In {\em 2018 {IEEE} International Conference on Image Processing,
  {ICIP} 2018, Athens, Greece, October 7-10, 2018}, pages 1363--1367. {IEEE},
  2018.

\bibitem{ribeiro2016why}
Marco~T{\'{u}}lio Ribeiro, Sameer Singh, and Carlos Guestrin.
\newblock "why should {I} trust you?": Explaining the predictions of any
  classifier.
\newblock In Balaji Krishnapuram, Mohak Shah, Alexander~J. Smola, Charu~C.
  Aggarwal, Dou Shen, and Rajeev Rastogi, editors, {\em Proceedings of the 22nd
  {ACM} {SIGKDD} International Conference on Knowledge Discovery and Data
  Mining, San Francisco, CA, USA, August 13-17, 2016}, pages 1135--1144. {ACM},
  2016.

\bibitem{rudin2021interpretable}
Cynthia Rudin, Chaofan Chen, Zhi Chen, Haiyang Huang, Lesia Semenova, and Chudi
  Zhong.
\newblock Interpretable machine learning: Fundamental principles and 10 grand
  challenges, 2021.

\bibitem{saxe2019mathematical}
Andrew~M. Saxe, James~L. McClelland, and Surya Ganguli.
\newblock A mathematical theory of semantic development in deep neural
  networks.
\newblock {\em Proceedings of the National Academy of Sciences},
  116(23):11537--11546, 2019.

\bibitem{sedghi2019singular}
Hanie Sedghi, Vineet Gupta, and Philip~M. Long.
\newblock The singular values of convolutional layers.
\newblock In {\em International Conference on Learning Representations}, 2019.

\bibitem{simonyan2015very}
Karen Simonyan and Andrew Zisserman.
\newblock Very deep convolutional networks for large-scale image recognition.
\newblock In Yoshua Bengio and Yann LeCun, editors, {\em 3rd International
  Conference on Learning Representations, {ICLR} 2015, San Diego, CA, USA, May
  7-9, 2015, Conference Track Proceedings}, 2015.

\bibitem{stewart1993early}
G.~W. Stewart.
\newblock On the early history of the singular value decomposition.
\newblock {\em {SIAM} Rev.}, 35(4):551--566, 1993.

\bibitem{szegedy2014intriguing}
Christian Szegedy, Wojciech Zaremba, Ilya Sutskever, Joan Bruna, Dumitru Erhan,
  Ian~J. Goodfellow, and Rob Fergus.
\newblock Intriguing properties of neural networks.
\newblock In Yoshua Bengio and Yann LeCun, editors, {\em 2nd International
  Conference on Learning Representations, {ICLR} 2014, Banff, AB, Canada, April
  14-16, 2014, Conference Track Proceedings}, 2014.

\bibitem{wei2015understanding}
Donglai Wei, Bolei Zhou, Antonio Torralba, and William~T. Freeman.
\newblock Understanding intra-class knowledge inside {CNN}, 2015.

\bibitem{wei2001ecg}
Jyh{-}Jong Wei, Chuang{-}Jan Chang, Nai{-}Kuan Chou, and Gwo{-}Jen Jan.
\newblock {ECG} data compression using truncated singular value decomposition.
\newblock {\em {IEEE} Trans. Inf. Technol. Biomed.}, 5(4):290--299, 2001.

\bibitem{xie2017aggregated}
Saining Xie, Ross~B. Girshick, Piotr Doll{\'{a}}r, Zhuowen Tu, and Kaiming He.
\newblock Aggregated residual transformations for deep neural networks.
\newblock In {\em 2017 {IEEE} Conference on Computer Vision and Pattern
  Recognition, {CVPR} 2017, Honolulu, HI, USA, July 21-26, 2017}, pages
  5987--5995. {IEEE} Computer Society, 2017.

\bibitem{xiong2019mixed}
Yunyang Xiong, Hyunwoo~J. Kim, and Vikas Singh.
\newblock Mixed effects neural networks (menets) with applications to gaze
  estimation.
\newblock In {\em {IEEE} Conference on Computer Vision and Pattern Recognition,
  {CVPR} 2019, Long Beach, CA, USA, June 16-20, 2019}, pages 7743--7752.
  Computer Vision Foundation / {IEEE}, 2019.

\bibitem{yoshida2017spectral}
Yuichi Yoshida and Takeru Miyato.
\newblock Spectral norm regularization for improving the generalizability of
  deep learning, 2017.

\bibitem{zhang2018interpreting}
Quanshi Zhang, Ruiming Cao, Feng Shi, Ying~Nian Wu, and Song{-}Chun Zhu.
\newblock Interpreting {CNN} knowledge via an explanatory graph.
\newblock In Sheila~A. McIlraith and Kilian~Q. Weinberger, editors, {\em
  Proceedings of the Thirty-Second {AAAI} Conference on Artificial
  Intelligence, (AAAI-18), the 30th innovative Applications of Artificial
  Intelligence (IAAI-18), and the 8th {AAAI} Symposium on Educational Advances
  in Artificial Intelligence (EAAI-18), New Orleans, Louisiana, USA, February
  2-7, 2018}, pages 4454--4463. {AAAI} Press, 2018.

\end{thebibliography}
